%% file: arxiv.tex
\documentclass[10pt, conference, compsocconf]{IEEEtran}

\usepackage{amsmath}
\usepackage{amssymb}
\usepackage{xspace}
\usepackage{array}

\usepackage{balance}

\DeclareMathOperator*{\argmax}{\mathrm{argmax}}
\DeclareMathOperator*{\argmin}{\mathrm{argmin}}
\newcommand{\eq}[1]{(eq.~\ref{#1})}

\newcommand{\todo}[1]{}
\newcommand{\xhdr}[1]{\subsection{#1}}

\usepackage{graphicx}

\renewcommand\refname{{\normalsize References}}

\begin{document}
\title{Learning Attitudes and Attributes from Multi-Aspect Reviews}

\author{Julian McAuley, Jure Leskovec, Dan Jurafsky\\ Stanford}

\newcommand{\modelname}{{\sc Pale Lager}\xspace}

\newcommand{\citet}[1]{\cite{#1}}
\newcommand{\citep}[1]{\cite{#1}}

\maketitle

\input{000abstract}

\IEEEpeerreviewmaketitle

\input{010intro}

\input{020data}

\input{030model}

\input{040learning}

\input{050morelearning}

\input{060experiments}

\balance

\input{080conclusion}

{\vspace{1mm}\noindent{{\bf Acknowledgements.}}} We thak \emph{oDesk}, and especially Paul Heymann, for their assistance and support in obtaining groundtruth labels. This research has been supported in part by NSF
IIS-1016909,
CNS-1010921,
CAREER IIS-1149837,
IIS-1159679,
Albert Yu \& Mary Bechmann Foundation,
Boeing,
Allyes,
Samsung,
Intel,
Alfred P. Sloan Fellowship and
the Microsoft Faculty Fellowship.

\end{document}

%% file: 000abstract.tex
\begin{abstract}
Most online reviews consist of plain-text feedback together with a single numeric score. However, there are multiple dimensions to products and opinions, and understanding the `aspects' that contribute to users' ratings may help us to better understand their individual preferences. For example, a user's impression of an audiobook presumably depends on aspects such as the \emph{story} and the \emph{narrator}, and knowing their opinions on these aspects may help us to recommend better products. In this paper, we build models for rating systems in which such dimensions are \emph{explicit}, in the sense that users leave separate ratings for each aspect of a product. By introducing new corpora consisting of five million reviews, rated with between three and six aspects, we evaluate our models on three prediction tasks: First, we uncover which parts of a review discuss which of the rated aspects. Second, we \emph{summarize} reviews, by finding the sentences that best explain a user's rating. Finally, since aspect ratings are \emph{optional} in many of the datasets we consider, we recover ratings that are missing from a user's evaluation. Our model matches state-of-the-art approaches on existing small-scale datasets, while scaling to the real-world datasets we introduce. Moreover, our model is able to `disentangle' content and sentiment words: we automatically learn content words that are indicative of a particular aspect as well as the aspect-specific sentiment words that are indicative of a particular rating.
\end{abstract}

%% file: 010intro.tex
\section{Introduction}

Online reviews, consisting of numeric ratings and plain-text feedback, are a valuable source of data for tasks such as product recommendation, summarization, and sentiment analysis. Making effective use of such reviews means understanding \emph{why} users evaluated products the way they did. Did a user dislike an audiobook because of the narrator or because of the story? If a user prefers toys of a certain brand, is it because that brand's toys are fun, or because they are educational? If a user describes a beer as having `grapefruit tones', what feature of the beer does this refer to, and are the words `grapefruit tones' praise or criticism?\footnote{This is an extended version of our ICDM paper \citep{icdm}.}

Naturally, users' opinions are \emph{multifaceted}, and answering such questions means understanding the different \emph{aspects} that contribute to their evaluation.
For example, consider the beer-rating website \emph{BeerAdvocate}, one of the datasets included in our study.
When a user evaluates a beer, their opinion is presumably influenced by the beer's look, smell, taste, and feel (palate). Furthermore, their opinions about such aspects may be conflicted: if a beer has a bad taste but a good palate, it might be described as having `stale hops, but a velvety body'; how can we learn that `body' refers to palate, `hops' refers to taste, and `stale' and `velvety' refer to negative and positive sentiments about those aspects?

To answer this, we consider product rating systems in which such aspects are \emph{explicit}, in the sense that reviews include \emph{multiple} ratings \citep{titov}, corresponding to different aspects of each product. For example, users of \emph{BeerAdvocate} provide ratings for each of the four sensory aspects mentioned above, in addition to their overall opinion. An example review from this corpus is shown in Figure \ref{fig:review}.

\begin{figure}
\vspace{2.5mm}
\small
\begin{center}
\fbox{\parbox{0.468\textwidth}{
\vspace{-1mm}
\begin{center}`Partridge in a Pear Tree', brewed by `The Bruery'\end{center}

Dark brown with a light tan head, minimal lace and low retention.
Excellent aroma of dark fruit, plum, raisin and red grape with light vanilla, oak, caramel and toffee.
Medium thick body with low carbonation.
Flavor has strong brown sugar and molasses from the start over bready yeast and a dark fruit and plum finish.
Minimal alcohol presence.
Actually, this is a nice quad.
\begin{center}
\begin{tabular}{ccccc}
Feel: 4.5 & Look: 4 & Smell: 4.5 & Taste: 4 & Overall: 4
\end{tabular}
\end{center}
\vspace{-2.1mm}
}}
\end{center}
\normalsize
\caption{An example review from \emph{BeerAdvocate} (of the beer `Partridge in a Pear Tree', a Californian Quadrupel). Each review consists of five ratings and free-text feedback. Our goal is to assign each of the sentences to one of the five aspects being evaluated (in this example the six sentences discuss look, smell, feel, taste, taste, and overall impression, respectively). \label{fig:review}}
\end{figure}

We consider three tasks on this type of data: First, can multi-aspect ratings be used as a form of weak supervision to learn language models capable of uncovering which sentences discuss each of the rated aspects? For example, using only multi-aspect rating data from many reviews (and no other labels), can we learn that `medium thick body with low carbonation' in Figure \ref{fig:review} refers to `feel'?
Moreover, can we learn that the word `warm' may be negative when describing the \emph{taste} of a beer, but positive when describing its \emph{color}? Second, can such a model be used to summarize reviews, which for us means choosing a subset of sentences from each review that best explain a user's rating? And third, since ratings for aspects are optional on many of the websites we consider, can missing ratings be recovered from users' overall opinions in addition to the review content?

Although sophisticated models have been proposed for this type of data \citet{brody,ganu,lu11,titov},
our primary goal is to design models that are \emph{scalable} and \emph{interpretable}. In terms of scalability, our models scale to corpora consisting of several million reviews. In terms of interpretability, while topic-modeling approaches
learn distributions of words used to describe each aspect \citep{brody,ganu,titov,titovMultigrain}, we separately model words that describe an aspect and words that describe \emph{sentiment about an aspect}; in this way we learn highly interpretable topic and sentiment lexicons simultaneously.

\xhdr{Present work}
We introduce a new model, which we name \emph{Preference and Attribute Learning from Labeled Groundtruth and Explicit Ratings}, or \modelname for short.
We introduce corpora of five million reviews from \emph{BeerAdvocate}, \emph{RateBeer}, \emph{Amazon}, and \emph{Audible}, each of which have been rated with between three and six aspects. \modelname can readily handle datasets of this size under a variety of training scenarios: in order to predict sentence aspects, the model can be trained with no supervision (i.e., using only aspect ratings), weak supervision (using a small number of manually-labeled sentences in addition to unlabeled data), or with full supervision (using only manually-labeled data). Using expert human annotators we obtain groundtruth labels for over ten thousand of the sentences in our corpora.

We find that while our model naturally benefits from increasing levels of supervision, our unsupervised model already obtains good performance on the tasks we consider, and produces highly interpretable aspect and sentiment lexicons. We discover that separating reviews into individual aspects is crucial when recovering missing ratings, as conflicting sentiments may appear in reviews where multiple aspects are discussed. However, we find that this \emph{alone} is not enough to recover missing ratings, and that to do so requires us to explicitly model relationships between aspects.

Our segmentation task is similar to those of \citet{brody,ganu,titov}, where the authors use aspect ratings to label and rank sentences in terms of the aspects they discuss. The same papers also discuss summarization, though in a different context from our own work; they define summarization in terms of ranking sentences across multiple reviews, whereas our goal is to choose sentences that best explain a user's multiple-aspect rating. More recently, the problem of recovering missing aspect ratings has been discussed in \citet{gupta10,lu11}.

Unlike topic modeling approaches, which learn word distributions over topics for each aspect, \modelname simultaneously models words that discuss an aspect and words that discuss the associated sentiment. 
For example, from reviews of the type shown in Figure \ref{fig:review}, we learn that nouns such as `body' describe the aspect `feel', and adjectives such as `thick' describe positive sentiment about that aspect, and we do so without manual intervention or domain knowledge. For such data we find that it is critical to separately model sentiment lexicons \emph{per-aspect}, due to the complex interplay between nouns and adjectives \citep{oldwine}.

\xhdr{Contributions}
We introduce a new dataset of approximately five million reviews with multi-aspect ratings, with over ten thousand manually annotated sentences. The data we introduce requires a model that can be trained on millions of reviews in a short period of time, for which we adapt high-throughput methods from computer vision. The models we propose produce highly interpretable lexicons of each aspect, and their associated sentiments.

The tasks we consider have been studied in \citet{brody,ganu,titov} (segmentation and ranking) and \citet{gupta10,lu11} (rating prediction). Although these previous approaches are highly sophisticated, they are limited to corpora of at most a few thousand reviews. We use \emph{complete} datasets (i.e., all existing reviews) from each of the sources we consider, and show that good performance can be obtained using much simpler models.

The main novelty of our approach lies in \emph{how} we model such data. For each aspect we separately model the words that discuss an aspect and words that discuss sentiment about an aspect. In contrast to topic models, this means that we learn sentiment-neutral lexicons of words that describe an aspect (which contain nouns like `head', `carbonation', and `flavor' in our \emph{BeerAdvocate} data), while simultaneously learning sentiment lexicons for each aspect (which contain adjectives like `watery', `skunky', and `metallic').

\xhdr{Further related work}
Reviews consisting of plain-text feedback and a single numeric score have proved a valuable source of data in many applications, including product recommendation \citep{netflix}, feature discovery \citep{popescu05}, review summarization \citep{hu04}, and sentiment analysis \citep{turney02}, among others. Understanding the multifaceted nature of such ratings has been demonstrated to lead to better performance at such tasks \citep{ganu,hu04,ng06}. Matrix factorization techniques \citep{koren} use nothing but numeric scores to learn latent `aspects' that best explain users' preferences, while clustering and topic-modeling approaches use nothing but review text to model `aspects' in terms of words that appear in document corpora \citep{blei,ErkanRadev04,Gamon05}. While such approaches may accurately model the data, the topics they learn are frequently not interpretable by users, nor are they representative of ratable aspects. This is precisely the issue addressed in \citet{titovMultigrain}, which attempts to uncover latent topics that are in some way similar to aspects on which users vote.

\citet{Snyder07multipleaspect} and \citet{titov} used multi-aspect ratings as a means to summarize review corpora.
As in our work, the authors of \citet{titov} use topic-models to identify `topics' whose words are highly correlated with the aspects on which users vote, and like us they apply their model to assign aspect labels to sentences in a review. Some more recent works that deal with multi-aspect rating systems are \citet{baccianella09,brody,ganu,lu09,lu11}. Finally, \citet{jo11}
examine whether similar models can be applied \emph{without} explicit aspect ratings, so that ratable aspects might be uncovered using only overall ratings.

A number of works model the relationship between \emph{aspects} and \emph{opinions on aspects} \citep{lin09,zhao10}. In \citet{popescu05} sentiment is associated with objective product features;
similarly, \citet{cameras} assigns sentiment labels to different product `facets' in a corpus of camera reviews, where users manually specify facets that are important to their evaluation. \citet{brody} and \citet{ganu} demonstrate that sentence aspects can be inferred from sentence sentiment, and they also introduce a publicly-available dataset which we include in our study.

Noting that aspect ratings are optional in many multi-aspect review systems, the problem of recovering missing ratings is discussed in \citet{gupta10}, and more recently in \citet{lu11}. First, we confirm their finding that multiple-aspect rating prediction depends on having separate sentiment models for each aspect. We then extend their work by explicitly modeling relationships between aspects.

Our work is also related to the discovery of \emph{sentiment lexicons} \citep{rao09,velikovich10}. Unlike topic-modeling approaches, whose `topics' are per-aspect word distributions, our goal is to separately model words that discuss an aspect and words that discuss sentiment about an aspect. Lexicon discovery is discussed in \citet{mohammad09}, and has been used for review summarization by \citet{blair08}, though such approaches require considerable manual intervention, and are not learned automatically. The use of language in review systems such as those we consider is discussed in \citet{oldwine} and \citet{Hatz98}, whose findings are consistent with the lexicons we learn.

\begin{table*}
\caption{Dataset statistics. \label{tab:datasets}}
\begin{center}
\begin{tabular}{llrrrr}
\multicolumn{1}{c}{\bf DATASET} & \multicolumn{1}{c}{\bf ASPECTS} & \multicolumn{1}{c}{\hspace{-2mm}\bf \#USERS} & \multicolumn{1}{c}{\bf \#ITEMS} & \multicolumn{1}{c}{\bf \#REVIEWS} & \multicolumn{1}{c}{\bf CC}\\
\hline
Beer (beeradvocate) & feel, look, smell, taste, overall & 33,387 & 66,051 & 1,586,259 & 0.64\\ 
Beer (ratebeer) & feel, look, smell, taste, overall & 40,213 & 110,419 & 2,924,127 & 0.66\\ 
Pubs (beeradvocate) & food, price, quality, selection, service, vibe & 10,492 & 8,763 & 18,350 & 0.29\\ 
Toys \& Games (amazon) & durability, educational, fun, overall & 79,994 & 267,004 & 373,974 & 0.65\\ 
Audio Books (audible) & author, narrator, overall & 7,009 & 7,004 & 10,989 & 0.74\\
\end{tabular}
\end{center}
\end{table*}

%% file: 020data.tex
\section{Datasets}
\label{sec:data}

The beer-rating websites {\emph{BeerAdvocate}} and {\emph{RateBeer}} allow users to rate beers using a five-aspect rating system. Ratings are given on four sensory aspects (feel, look, smell, and taste), in addition to an overall rating.
From \emph{BeerAdvocate} we also obtain reviews of {\emph{Pubs}}, which are rated in terms of food, price, quality, selection, service, and vibe.

All \emph{Amazon} product reviews allow users to rate items in terms of their overall quality. The {\emph{Toys \& Games}} category allows users to provide further feedback, by rating products in terms of fun, durability, and educational value.

Finally, the audiobook rating website {\emph{Audible}} allows users to rate audiobooks in terms of the author and the narrator, in addition to their overall rating.

These datasets are summarized in Table \ref{tab:datasets}. The `CC' column shows the average correlation coefficient across all pairs of aspects. Our \emph{Pubs} data has the lowest correlation between aspects, and in fact some aspects are negatively correlated (price is negatively correlated with both food and service, as we would expect for pub data).

We briefly mention a variety of other websites that provide similar rating systems, including \emph{TheBeerSpot}, \emph{BeerPal}, \emph{Yahoo!~Hotels}, \emph{Yahoo!~Things to Do}, \emph{TigerDirect}, \emph{BizRate}, \emph{TripAdvisor}, and \emph{DP-Review}, among others.

We also obtained the \emph{CitySearch} dataset used in \citet{brody,ganu}. This dataset consists of 652 reviews, which are labeled using four aspects (food, ambiance, price, and staff). This data differs from our own in the sense that aspects are not determined from rating data (\emph{CitySearch} includes only overall ratings), but rather aspects and user sentiment are determined by human annotators. Sentiment labels are \emph{per sentence} (rather than per review), so to use their data with our method, we treat their sentiment labels (positive, negative, neutral, and conflicted) as four different ratings. We adapt our method so that ratings are indexed \emph{per sentence} rather than \emph{per aspect}, though we omit details for brevity.

\subsection{Groundtruth labels}
\label{sec:groundtruth}

Since we wish to learn which aspects are discussed in each sentence of a review, to evaluate our method we require groundtruth labels for a subset of our data. Our first author manually labeled 100 reviews from each of our datasets, corresponding to 4,324 sentences in total. Labels for each sentence consist of a single aspect, in addition to an `ambiguous/irrelevant' label.

For our \emph{BeerAdvocate} data, we obtained additional annotations using crowdsourcing. Using Amazon's Mechanical Turk, we obtained labels for 1,000 reviews, corresponding to 9,245 sentences. To assess the quality of these labels, we computed Cohen's kappa coefficient (to be described in Section \ref{sec:supervised}), a standard measure of agreement between two annotators. Unfortunately, the Mechanical Turk annotations agreed with our own labels in only about 30\% of cases, corresponding to $\kappa=0.11$, which is not significantly better than a random annotator (the 4\% of sentences labeled as `ambiguous' were not used for evaluation).

To address this, we used the crowdsourcing service \emph{oDesk}, which allows requesters to recruit individual workers with specific skills. We recruited two `expert' beer-labelers based on their ability to answer some simple questions about beer.
Both annotators labeled the same 1,000 reviews independently, requiring approximately 40 hours of work. These experts agreed with a kappa score of 0.93, and obtained similar scores against the 100 reviews labeled by our first author (who is also a beer expert).

Since \emph{RateBeer} reviews are similar to those of \emph{BeerAdvocate}, rather than annotating reviews from both corpora, from \emph{RateBeer} we obtained annotations from non-English reviews (which are far more common in the \emph{RateBeer} corpus). We identified 1,295 Spanish and 19,998 French reviews, and annotated 742 sentences from the two corpora, again with the help of expert labelers.

Code, data, and groundtruth labels shall be released at publication time.

%% file: 030model.tex
\section{The \modelname Model}

\modelname models aspects, and ratings on aspects, as a function of the words that appear in each sentence of a review. Our goal is to simultaneously learn which words discuss a particular aspect, and which words are associated with a particular rating. For example, in our \emph{BeerAdvocate} data, the word `flavor' might be used to discuss the `taste' aspect, whereas the word `amazing' might indicate a 5-star rating. Thus if the words `amazing flavor' appear in a sentence, we would expect that the sentence discusses `taste', and that the `taste' aspect has a high rating. As a first approximation, nouns can be thought of as `aspect' words, and adjectives as `sentiment' words; we find that this intuition closely matches the parameters we learn.

We first introduce the notation used throughout the paper.
Suppose our review corpus $(\mathcal R, \mathcal V)$ consists of \emph{reviews} $\mathcal R = \lbrace r_1 \ldots r_R \rbrace$ and \emph{ratings} $\mathcal V = \lbrace v_1 \ldots v_R \rbrace$. Next assume that each review $r_i$ is divided into \emph{sentences} $s \in r_i$, and that each rating $v_i$ is divided into $K$ \emph{aspects} $\lbrace v_{i1} \ldots v_{iK}\rbrace$ (e.g.~ratings on smell, taste, overall impression, etc.). Finally, assume that each sentence is further divided into \emph{words}, $w \in r_{is}$.

We assume that each sentence in a review discusses a single aspect; alternately we could model one aspect per \emph{word} or per \emph{paragraph}, though one aspect per sentence matches what appears in existing work \citep{brody,ganu,titov}.

Our goal is to differentiate words that discuss an aspect from words that discuss the associated sentiment. To do so, we separate our model into two parameter vectors, $\theta$ and $\phi$, which respectively encode these two properties.
In our model, the probability that a sentence $s$ discusses a particular aspect $k$, given the ratings $v$ associated with the review, is
\begin{multline}
 P^{(\theta, \phi)}(\text{aspect}(s) = k\ |\ \text{sentence\ } s, \text{rating\ } v) =\\ \frac{1}{Z^{(\theta, \phi)}_s} \exp \sum_{w\in s} \Bigl\lbrace \!\! \underbrace{\vphantom{\phi_{kv_kw}}\theta_{kw}}_{\text{aspect weights}} + \underbrace{\phi_{kv_kw}}_{\text{sentiment weights}} \!\! \Bigr\rbrace.
\label{eq:probtopic}
\end{multline}
The normalization constant $Z_s$ is
\begin{equation}
 Z^{(\theta, \phi)}_s = \sum_{k = 1}^K \exp \sum_{w\in s} \Bigl\lbrace \theta_{kw} + \phi_{kv_kw} \Bigr\rbrace.
\end{equation}
Note that $\theta_k$ is indexed by the aspect $k$, so that we learn which words are associated with each of the $K$ aspects. Alternately, $\phi_{kv_k}$ is indexed by the aspect $k$, \emph{and the rating for that aspect} $v_k$; this way, for each aspect we learn which words are associated with each star rating. While using $\theta$ and $\phi$ together has no more expressive power than using $\phi$ alone, we find that separating the model in this way is critical for interpretability. Another option would be to have a single sentiment parameter $\phi_{v_k}$ for all aspects; however, we find that each aspect uses different sentiment words (e.g.~`delicious' for taste, `skunky' for smell), so it is beneficial to learn sentiment models per-aspect \citep{oldwine}.

Assuming that aspects for each sentence are chosen independently, we can write down the probability for an entire review (and an entire \emph{corpus}) as
\begin{equation}
 p^{(\theta, \phi)}(\text{aspects} | \mathcal R, \mathcal V) = \prod_{i = 1}^R \prod_{s \in r_i} p^{(\theta, \phi)}(\text{aspect}(s) | s, v_i).
\label{eq:corpusprob}
\end{equation}
We will now show how to learn aspect labels and parameters so as to maximize this expression.

%% file: 040learning.tex
\section{Learning}
\label{sec:learning}

We describe three learning schemes, which use increasing levels of supervision in the form of sentence labels.
As we show in Section \ref{sec:experiments}, increased supervision leads to higher accuracy, though even without supervision we can obtain good performance given enough data.

\subsection{Unsupervised Learning}

\begin{figure*}
 \includegraphics[width=\textwidth]{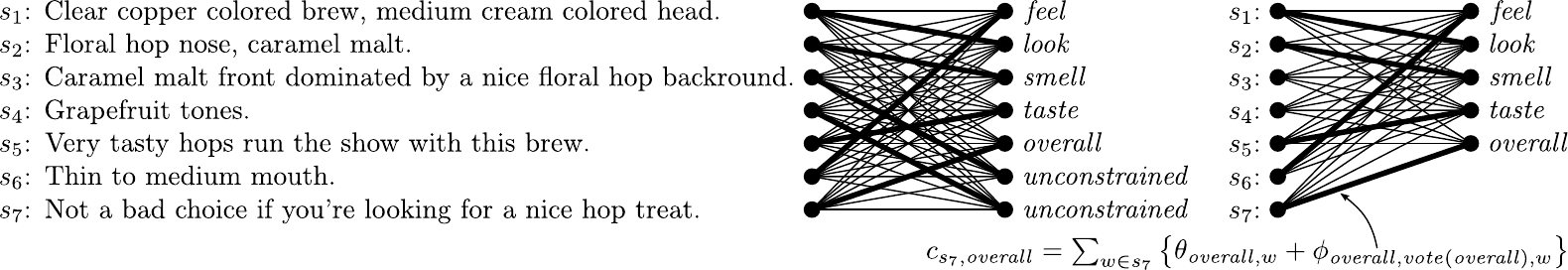}
\vspace{-6mm}
\caption{Both our segmentation and summarization tasks can be expressed as weighted bipartite graph cover. Each of the sentences at left (from a \emph{BeerAdvocate} review) must be matched to an aspect. The optimal cover is highlighted using bold edges. For the segmentation task (left graph), five nodes are constrained to match to each of the five aspects, ensuring that each aspect appears at least once in the segmentation (the remaining two unconstrained aspects are both `smell' in this case). The summarization task (right graph) includes precisely one node for each aspect, so that each aspect is summarized using the sentence that most closely aligns with that aspect's rating. \label{fig:bipartite}}
\end{figure*}

Unsupervised learning proceeds by choosing the parameters $(\hat{\theta}, \hat{\phi})$ and the latent aspect assignments $\hat{t}$ so as to maximize the log-likelihood of the corpus:
\begin{equation}
 (\hat{\theta}, \hat{\phi}), \hat{t} = \argmax_{(\theta, \phi), t} \underbrace{\log p^{(\theta, \phi)}(t | \mathcal R, \mathcal V)}_{\text{corpus probability}} - \underbrace{\Omega(\theta, \phi)}_{\text{regularizer}}.
\label{eq:opt_problem_unsupervised}
\end{equation}

Optimization proceeds by coordinate ascent on $(\theta, \phi)$ and $t$, i.e., by alternately optimizing
\begin{eqnarray}
 t^{i} & = & \argmax_{t} \log p^{(\theta, \phi)^i} (t | \mathcal R, \mathcal V) \label{eq:opt1}\\
 (\theta, \phi)^{t+1} & = & \argmax_{(\theta, \phi)} \log p^{(\theta, \phi)}(t^i | \mathcal R, \mathcal V) - \Omega(\theta, \phi) \label{eq:opt2}
\end{eqnarray}
until convergence, i.e., until $t^{i} = t^{i-1}$. Optimizing \eq{eq:opt1} merely consists of maximizing \eq{eq:probtopic} independently for each sentence. Noting that the model is concave in $(\theta, \phi)$, optimization of \eq{eq:opt2} proceeds by gradient ascent, where partial derivatives can be easily calculated. We regularize using the squared $\ell_2$ norm,
 $\Omega(\theta, \phi) = ||\theta||_2^2 + ||\phi||_2^2$.

Being a local optimization procedure, coordinate ascent is sensitive to initialization. We initialize $\theta$ by setting $\theta_{k,k} = 1$ for each aspect $k$ (e.g.~$\theta_{\mathit{taste},\mathit{`taste'}} = 1$). In practice this means that initially a sentence is assigned to an aspect if it explicitly mentions the name of that aspect. Other parameters were initialized randomly; we selected the model with the highest log-likelihood among 64 random restarts.

Finally, we note that \eq{eq:probtopic} is underconstrained, in the sense that adding a constant to $\theta_{kw}$ and subtracting the same constant from $\phi_{k\cdot w}$ has no effect on the model. To address this, we add an additional constraint that
\begin{equation}
 \sum_{v} \phi_{kvw} = 1, \text{\ for all $k$, $w$}.
\end{equation}
This has no effect on performance, but in our experience leads to significantly more interpretable parameters.

\subsection{Enforcing Diversity in the Predicted Output}
\label{sec:diversity}

An issue we encountered with the above approach is that `similar' aspects tended to coalesce, so that sentences from different aspects were erroneously assigned a single label. For example, on \emph{BeerAdvocate} data, we noticed that `smell' and `taste' words would often combine to form a single aspect.
From the perspective of the regularizer, this makes perfect sense: ratings for `smell' and `taste' are highly correlated,
and similar words are used to describe both; 
thus setting one aspect's parameter vector to zero significantly reduces the regularization cost, while reducing the log-likelihood only slightly.

To address this, we need to somehow enforce \emph{diversity} in our predictions, that is, we need to encode our knowledge that all aspects should be discussed. In practice, we enforce such a constraint \emph{per-review}, so that we choose the most likely assignments of aspects, subject to the constraint that each aspect is discussed at least once.
We find this type of constraint in computer vision applications: for example, to localize a pedestrian, we might encode the fact that each of their limbs must appear exactly once in an image \citep{lgm}. Instead of matching image coordinates to limbs, we match sentences to aspects, but otherwise the technology is the same. In \citet{lgm}, such a constraint is expressed as \emph{bipartite graph cover}, and is optimized using \emph{linear assignment}.

We construct a bipartite graph for each review $r$, which matches $|r|$ sentences to $|r|$ aspects. From \eq{eq:probtopic} we define the compatibility between a sentence $s$ and aspect $k$:
\begin{equation}
 c_{sk} = \sum_{w \in s} \lbrace \theta_{kw} + \phi_{kv_kw}\rbrace.
\label{eq:compatibility}
\end{equation}
Next we define edge weights in terms of this compatibility function. Noting that each of the $K$ aspects \emph{must} be included in the cover, our weight matrix $A^{(r)}$ is defined as
\begin{equation}
 A^{(r)}_{s,l} = \biggl\lbrace\hspace{-7mm} \overbrace{ \underbrace{ \begin{array}{ll} c_{sl} & \text{if $1 \leq l \leq K$}\\
                                          \max_{k} c_{r_sk} & \text{otherwise}
                        \end{array} }_{\text{other sentences can match any aspect}} }^{\text{each of the $K$ aspects must have a matching sentence}}\hspace{-7mm},
\end{equation}
and the optimal cover is given by
\begin{equation}
 \hat{f} = \argmax_{f} \sum_{s\in r} A^{(r)}_{s,f(s)},
\label{eq:lap}
\end{equation}
which is found using the Kuhn-Munkres algorithm.

This entire procedure is demonstrated in Figure \ref{fig:bipartite}, where the assignment matrix $A^{(r)}$ is visualized using a weighted bipartite graph, so that $\hat{f}$ becomes a cover of that graph. The nodes on the left of the graph correspond to the sentences in the review, while the nodes on the right correspond to their assignments. $K$ of the nodes on the right are constrained to match each of the $K$ aspects, while the remaining nodes may match to any aspect.

The same bipartite matching objective can also be used for our summarization task.
Here, our goal is to predict for each aspect the sentence that best explains that aspect's rating. Using the compatibility function of \eq{eq:compatibility}, our goal is now to choose the $K$ sentences that are \emph{most compatible} with the $K$ aspects. This idea is depicted on the right of Figure \ref{fig:bipartite}. These constraints are discarded for reviews with fewer than $K$ sentences.

If the hard constraint that every aspect must be discussed proves too strong, it can be relaxed by adding additional `unconstrained' nodes: for example, adding two additional nodes would mean that each review must discuss at least $K - 2$ unique aspects. Alternately, in datasets where aspects are easily separable (such as \emph{CitySearch}), this constraint can be discarded altogether, or discarded at test time. However, for datasets whose aspects are difficult to distinguish (such as \emph{BeerAdvocate}), this constraint proved absolutely critical.

\subsection{Semi-Supervised Learning}

The semi-supervised variant of our algorithm is no different from the unsupervised version, except that the probability is conditioned on some fraction of our groundtruth labels $t'$, i.e., our optimization problem becomes
\begin{equation}
 (\hat{\theta}, \hat{\phi}), \hat{t} = \argmax_{(\theta, \phi), t} \underbrace{\log p^{(\theta, \phi)}(t | \mathcal R, \mathcal V, t')}_{\text{corpus probability}} - \underbrace{\Omega(\theta, \phi)}_{\text{regularizer}}.
\label{eq:opt_problem_weakly_supervised}
\end{equation}
In addition, we initialize the parameters $\theta$ and $\phi$ so as to maximize the likelihood of the observed data $t'$.

\subsection{Fully-Supervised Learning}
\label{sec:supervised}

Given fully-labeled data, it would be trivial to choose $\hat{\theta}$ and $\hat{\phi}$ so as to maximize the log-likelihood of \eq{eq:opt_problem_unsupervised}. However, a more desirable option is to learn parameters so as to directly optimize the criterion used for evaluation.

Cohen's kappa statistic is a standard accuracy measure for document labeling tasks \citep{cohen}. It compares an annotator or algorithm's performance to that of a random annotator:
\begin{equation}
 \kappa(a,b) = \frac{P(a \text{\ agrees with\ } b) - 1/K}{1 - 1/K}.
\label{eq:kappa}
\end{equation}
$\kappa = 0$ corresponds to random labeling, $0 < \kappa \leq 1$ corresponds to some level of agreement, while $\kappa < 0$ corresponds to disagreement. If two annotators $a$ and $b$ label a corpus with aspects $t^{(a)}$ and $t^{(b)}$, then
\begin{equation}
 P(a \text{\ agrees with\ } b) = 1 - \Delta_{0/1}(t^{(a)}, t^{(b)}),
\label{eq:agreement}
\end{equation}
where $\Delta_{0/1}$ is the $0/1$ loss.
Critically, since kappa is a monotonic function of the $0/1$ loss, a predictor trained to minimize the $0/1$ loss will maximize Cohen's kappa statistic. We train a predictor based on the principle of regularized risk minimization, i.e., we optimize
\begin{equation}
 \hat{\theta}, \hat{\phi} = \argmin_{\theta, \phi} \underbrace{\Delta_{0/1}(t^{(\theta, \phi)}, t')}_{\text{empirical risk}} + \underbrace{\vphantom{\Delta_{0/1}(t^{(\theta, \phi)}, t')}\Omega(\theta,\phi)}_{\text{regularizer}},
\label{eq:opt_problem}
\end{equation}
so that $\hat{\theta}$ and $\hat{\phi}$ are chosen so as to minimize the $0/1$ loss on some training data $t'$ provided by an annotator.

If not for the diversity constraint of Section \ref{sec:diversity}, optimization of \eq{eq:opt_problem} would be independent for each sentence, and could be addressed using a multiclass SVM or similar technique. However, the diversity constraint introduces \emph{structure} into the problem so that predictions cannot be made independently. Thus we require an optimization technique designed for structured output spaces, such as that of \citep{tsoch05}. The use of bipartite graph cover objectives in structured learning is addressed in \citet{lgm}, where an objective similar to that of \eq{eq:lap} is used to match keypoints in images.
We adapt their framework to our problem,
which can be shown to minimize a convex upper bound on \eq{eq:opt_problem}.

%% file: 050morelearning.tex
\section{Learning to Predict Ratings from Text}
\label{sec:votepredict}

In many websites with multiple aspect ratings, ratings for aspects are \emph{optional}, while only `overall' ratings are mandatory. For example, our 10,989 \emph{Audible} reviews represent only those where all three aspects (author, narrator, overall) were rated. In total there were 199,810 reviews in our crawl that included an overall vote but were missing an aspect rating.
Predicting such missing ratings may help us to understand \emph{why} users voted the way they did. We will learn models for this task from users who entered complete ratings.

A na\"ive solution would be to learn parameters $\gamma_{kv_kw}$ for each aspect $k$ and rating $v_k$, using fully-rated reviews as training data. That is, each rating $v_{ik}$ for review $r_i$ and aspect $k$ would be predicted according to
\begin{equation}
 v^{(\gamma)}_{ik} = \argmax_{v} \sum_{w \in r_i} \gamma_{kvw}.
\label{eq:predict_unsegmented}
\end{equation}
We shall see in Section \ref{sec:experiments} that this proves ineffective when users have mixed feelings about different aspects: both positive and negative words appear together in reviews, making it difficult to `tease-apart' users' opinions.

An appealing solution to this problem consists of using \emph{segmented} text
to predict ratings for each aspect, i.e.,
\begin{equation}
 v^{(\gamma)}_{ik} = \argmax_{v} \!\!\!\!\!\!\! \underbrace{\sum_{s \in r_i} \delta(\hat{t}_{is} = k)}_{\text{sentences labeled with aspect $k$}} \!\!\!\!\!\!\! \sum_{w \in s} \gamma_{kvw}.
\label{eq:predict_segmented}
\end{equation}
However, we found that this approach \emph{also} performs poorly, even when highly accurate sentence labels are available \citet{lu11}. A simple explanation is that different aspects are highly correlated: for example, when learning na\"ive predictors from unsegmented text on \emph{BeerAdvocate} data as described in \eq{eq:predict_unsegmented}, we found that the word `skunky' was among the strongest 1-star predictors \emph{for all aspects}, even though the word clearly refers only to smell. Not surprisingly, a product that smells `skunky' is unlikely to be rated favorably in terms of its taste; by predicting ratings from segmented text as in \eq{eq:predict_segmented}, we fail to exploit this correlation.

\begin{figure}
\begin{center}
 \includegraphics[scale=1]{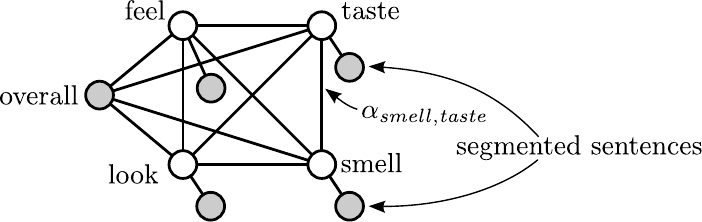}
\caption{Graphical model for predicting ratings from segmented text. Grey nodes represent observations, whereas white nodes represent variables. The model incorporates both segmented text and relationships between aspects. \label{fig:gm_votes}}
\end{center}
\end{figure}

Instead, the model we propose uses segmented text, but explicitly encodes relationships between aspects. Our model is depicted in Figure \ref{fig:gm_votes}. In addition to conditioning on segmented text, the `smoothness' term $\alpha$ encodes how likely two ratings are to co-occur for different aspects:
\begin{equation}
 v^{(\gamma,\alpha)}_{i} = \argmax_{v} \sum_{k} \sum_{s \in r_i} \delta(t_{is} = k) \sum_{w \in s} \!\gamma_{kv_kw} +
 \sum_{i \neq j} \alpha_{ijv_iv_j}.
\label{eq:predict_graphical}
\end{equation}
For example, $\alpha_{\mathit{smell}, \mathit{taste}, 1, 5}$ encodes the penalty for a 1-star `smell' vote to co-occur with a 5-star `taste' vote; in practice $\alpha$ prevents such an unlikely possibility from occurring.

We train each of the above predictors (eqs.~\ref{eq:predict_unsegmented}, \ref{eq:predict_segmented}, and \ref{eq:predict_graphical}) so as to minimize the $\ell_2$ error of the prediction compared to the groundtruth ratings used for training, i.e.,
\begin{equation*}
 (\hat{\gamma},\hat{\alpha}) = \argmin_{\gamma,\alpha} \sum_{i = 1}^{R'} \sum_{k \neq \mathit{overall}} || v^{(\gamma,\alpha)}_{ik} - v_{ik} ||_2 + \Omega(\gamma, \alpha).
\end{equation*}
We optimize this objective using a multiclass SVM in the case of (eqs.~\ref{eq:predict_unsegmented} and \ref{eq:predict_segmented}), though for \eq{eq:predict_graphical} the term $\alpha$ introduces dependencies between ratings, so we again use structured learning techniques as in Section \ref{sec:supervised} \citep{tsoch05}.

%% file: 060experiments.tex
\section{Experiments}
\label{sec:experiments}

We evaluate \modelname on our segmentation, summarization, and rating prediction tasks. Segmentation requires us to predict aspect labels for each sentence in our review corpora, while summarization requires us to choose one sentence per aspect for each review. For the first two tasks we report the accuracy (i.e., the fraction of correct predictions), which is related to Cohen's kappa by \eq{eq:kappa}. For rating prediction we report the $\ell_2$ error of the predicted ratings, after scaling ratings to be in the range $[0,1]$. Even in the largest experiments we report, \modelname could be trained in a few hours using commodity hardware.

We randomly split groundtruth data from each of our corpora into training and test sets. Our \emph{unsupervised} algorithm uses entire corpora, but ignores groundtruth sentence labels; our \emph{semi-supervised} algorithm also uses entire corpora, and conditions on the labeled training data; our \emph{fully-supervised} algorithm uses \emph{only} the labeled training data. All algorithms are evaluated on groundtruth labels from the test set. For our rating prediction task, we further split our unlabeled data into training and test sets, so that our segmentation algorithms are not trained using the ratings we are trying to predict.

\todo{Baselines, probably with multiclass SVMs and and some variant of LDA.}

\begin{figure*}
\begin{center}
 \includegraphics[width=0.9\textwidth]{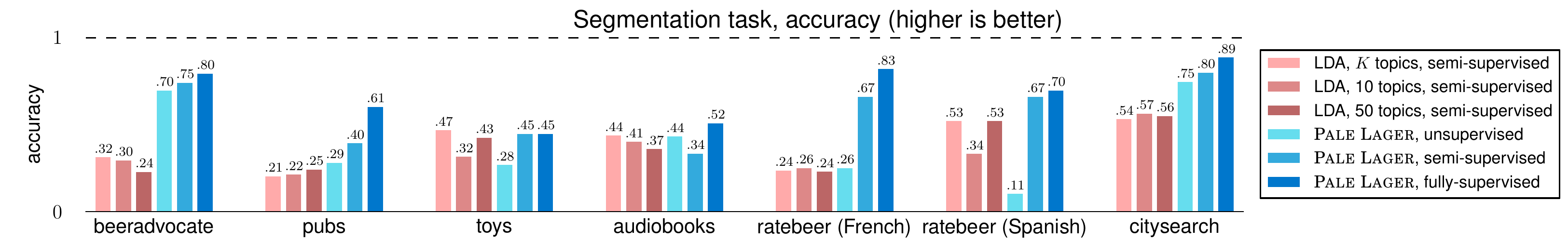}

\vspace{2mm}
 \includegraphics[width=0.9\textwidth]{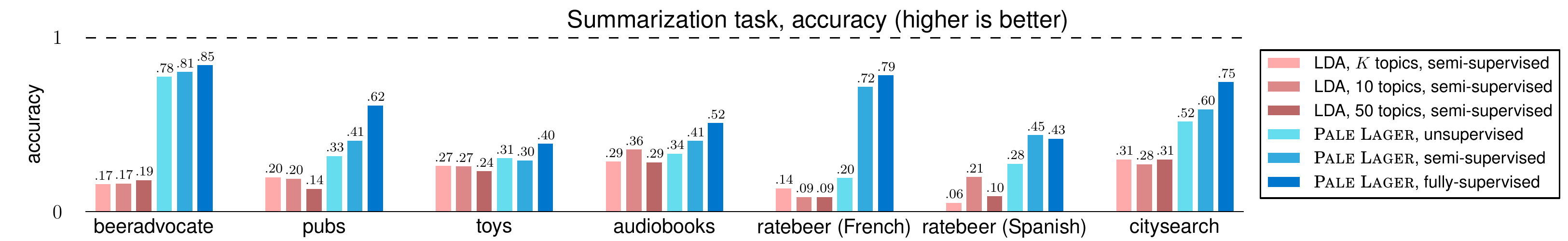}

\vspace{2mm}
 \includegraphics[width=0.9\textwidth]{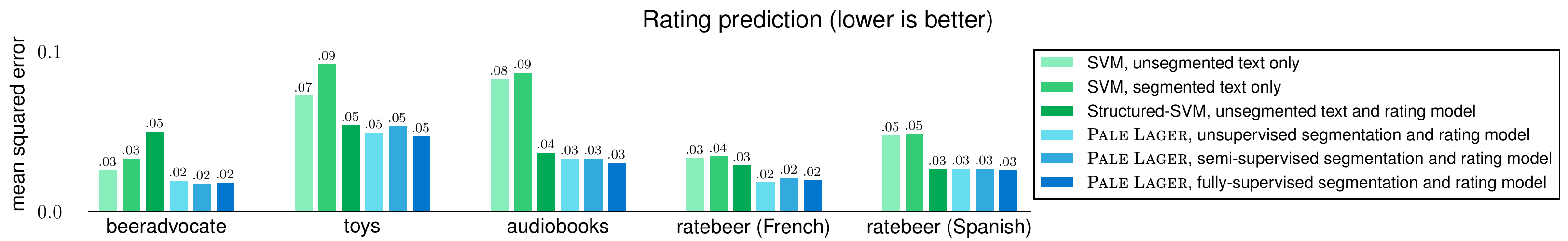}
\end{center}
\vspace{-3mm}
\caption{Performance of \modelname and baselines on our segmentation task (top), our summarization task (middle), and our rating prediction task (bottom). Results are shown in terms of accuracy (higher is better) and mean squared error (lower is better). \label{fig:perf}}
\end{figure*}

\subsection{Review Segmentation}

Figure \ref{fig:perf} (top) shows the performance of \modelname on the seven datasets we consider. As expected, semi-supervised learning improves upon unsupervised learning (by 45\% on average), and fully-supervised learning outperforms semi-supervised learning by a further 17\%; the sole exception occurs on \emph{Audible} data, which is possibly due to overfitting. Despite the good performance of our unsupervised method on \emph{BeerAdvocate} data, it performs poorly on non-English \emph{RateBeer} data. The simplest explanation is merely the paucity of non-English data, revealing that while this task can be approached without supervision, it requires many reviews to do so (though this could be addressed using seed-words). Once we add supervision, we observe similar performance across all three beer datasets.

As a baseline we compare \modelname to \emph{Latent Dirichlet Allocation} \citep{blei}. We train LDA with different numbers topics, and use our training labels to identify the optimal correspondence between topics and aspects (so in this sense the process is \emph{semi-supervised}). Our semi-supervised model outperforms this baseline in 5 out of 7 cases and by 48\% on average; two exceptions occur in datasets where users tend to focus on their overall evaluation and do not discuss aspects (e.g.~toy reviews rarely discuss durability or educational value). We acknowledge the existence of more sophisticated variants of LDA, though we are not aware of suitable alternatives that scale to millions of reviews; we used \emph{Online LDA} as implement in \emph{Vowpal Wabbit} \citep{hoffman}, which required a few hours to train on our largest dataset.

In all experiments fully-supervised learning outperforms semi-supervised learning, even though the semi-supervised algorithm has access to both labeled \emph{and} unlabeled data. An explanation is that our semi-supervised algorithm optimizes the log-likelihood, while the fully-supervised algorithm directly optimizes the accuracy score used for evaluation.
It is certainly possible that by using latent-variable structured learning techniques our fully-supervised algorithm could be extended to make use of unlabeled data \citep{latentSVM}.

\begin{table}
\caption{CitySearch results, using accuracy scores from \citet{lu11}. \label{tab:citysearch}}
\begin{center}
 \begin{tabular}{|lr|}
  \hline
    Always label as `food' & 0.595\\
    LDA \citep{blei} & 0.477\\
    MultiGrain LDA \citep{titovMultigrain} & 0.760\\
    Segmented Topic Models \citep{du10} & 0.794\\
    Local LDA \citep{brody} & 0.803\\
    Support Vector Machine & 0.830\\
\hline
    \modelname, unsupervised & 0.751\\
    \modelname, semi-supervised & 0.805\\
    \modelname, fully-supervised & 0.892\\
  \hline
 \end{tabular}
\end{center}
\end{table}

\begin{figure*}
\includegraphics[width=0.2\textwidth]{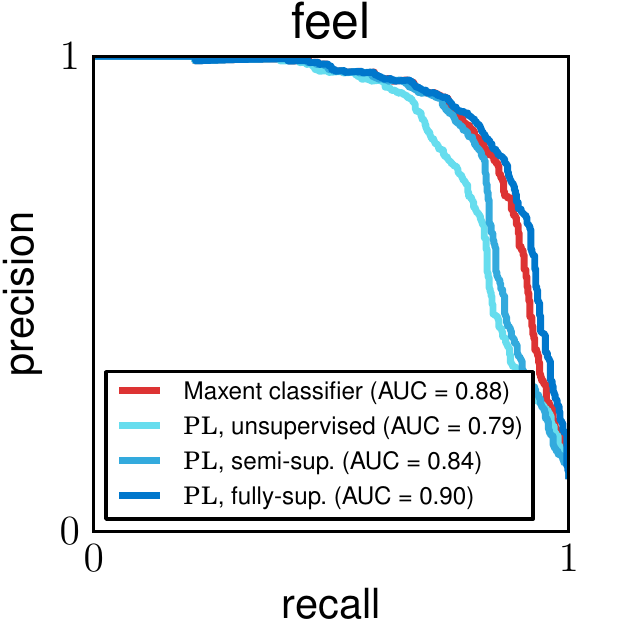}%
\includegraphics[width=0.2\textwidth]{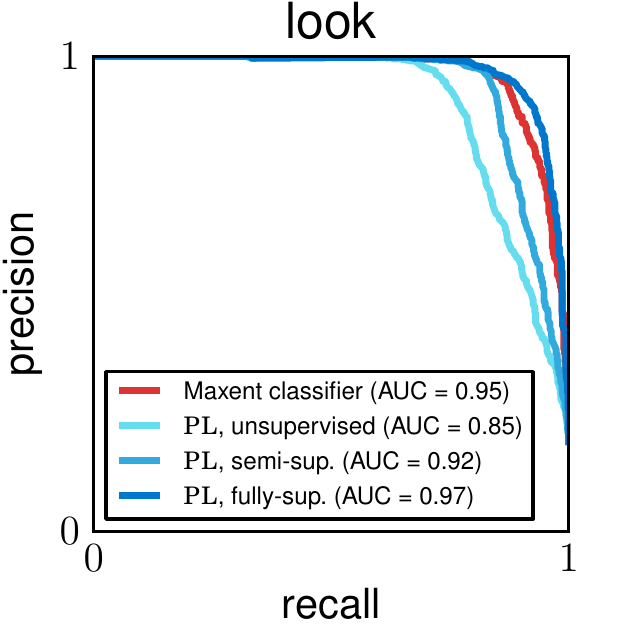}%
\includegraphics[width=0.2\textwidth]{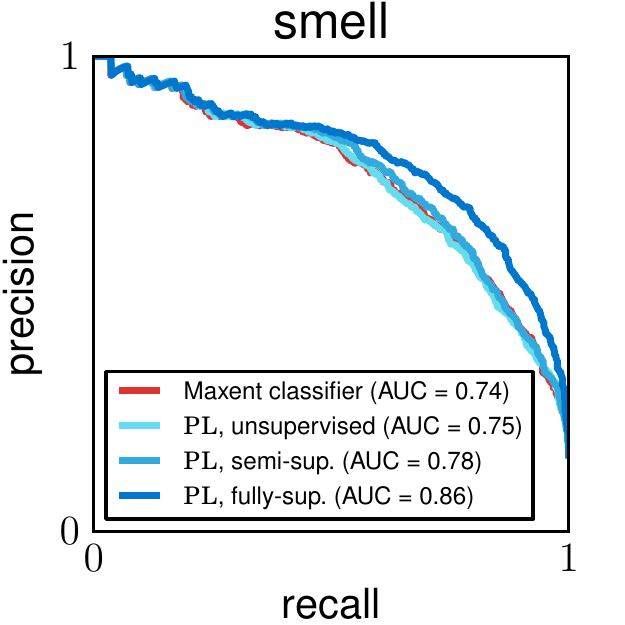}%
\includegraphics[width=0.2\textwidth]{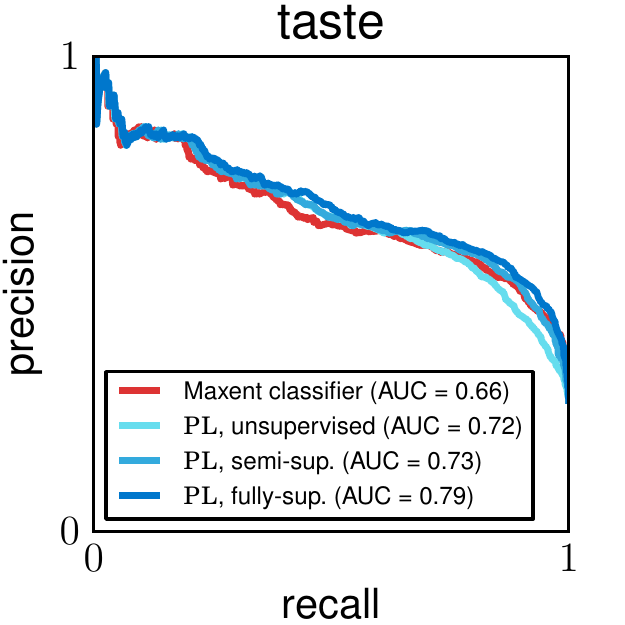}%
\includegraphics[width=0.2\textwidth]{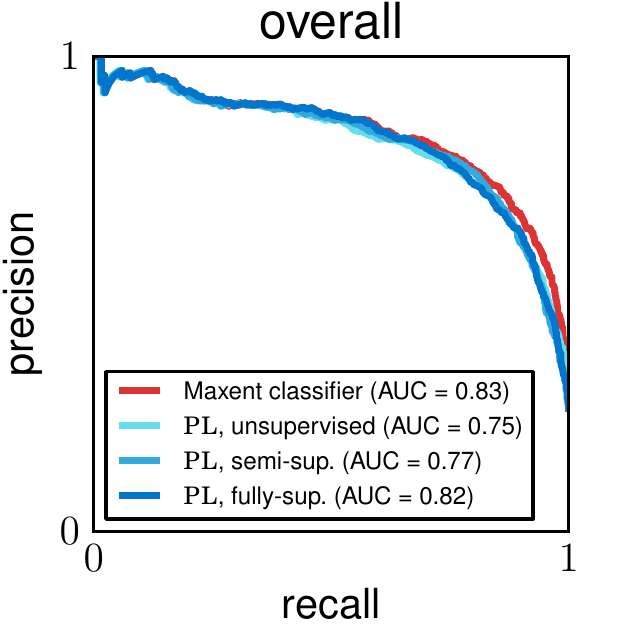}
\vspace{-5mm}
\caption{Precision recall curves for sentence ranking (blue curves are the \modelname model). Like in \citet{titov}, we find that our unsupervised method achieves performance close to that of a fully-supervised Maximum Entropy classifier. However, we also report that our semi-supervised method matches the performance of the maxent classifier, and our fully supervised method outperforms maximum entropy classification significantly. In terms of Mean Average Precision, Maxent = 0.82 (fully-supervised), \modelname = 0.76 (unsupervised), 0.81 (semi-supervised), 0.87 (fully-supervised). \label{fig:prcurves}}
\end{figure*}

\subsubsection{Performance on CitySearch data}

To our knowledge, the 652 review \emph{CitySearch} dataset from \citet{brody,ganu} is the only publicly-available dataset for the aspect labeling task we consider. In Table \ref{tab:citysearch} we report published results from \citet{lu11}. For comparison we used the same seed-words as in their study, which aid performance significantly.
Note that in this dataset supervision takes the form of \emph{per-sentence} ratings, rather than \emph{per-aspect} ratings, though our method can be adapted to handle both.
\modelname is competitive with highly sophisticated alternatives, while requiring only a few seconds for training. The supervised version of our model, which jointly models text and ratings, outperforms (by 7\%) an SVM that uses text data alone. Overall, this is a promising result: our method is competitive with sophisticated alternatives on a small dataset, and scales to the real-world datasets we consider.

\begin{figure*}
\begin{tabular}{m{0.08\textwidth}m{0.26\textwidth}m{0.26\textwidth}m{0.26\textwidth}}
Aspect $k$ & \ \ \ \ \ \ Aspect words $\theta_k$ & \ \ 2-star sentiment words $\phi_{k,2}$ & \ \ 5-star sentiment words $\phi_{k,5}$\\[-3pt]
Feel & {\includegraphics[angle=-90,scale=0.2]{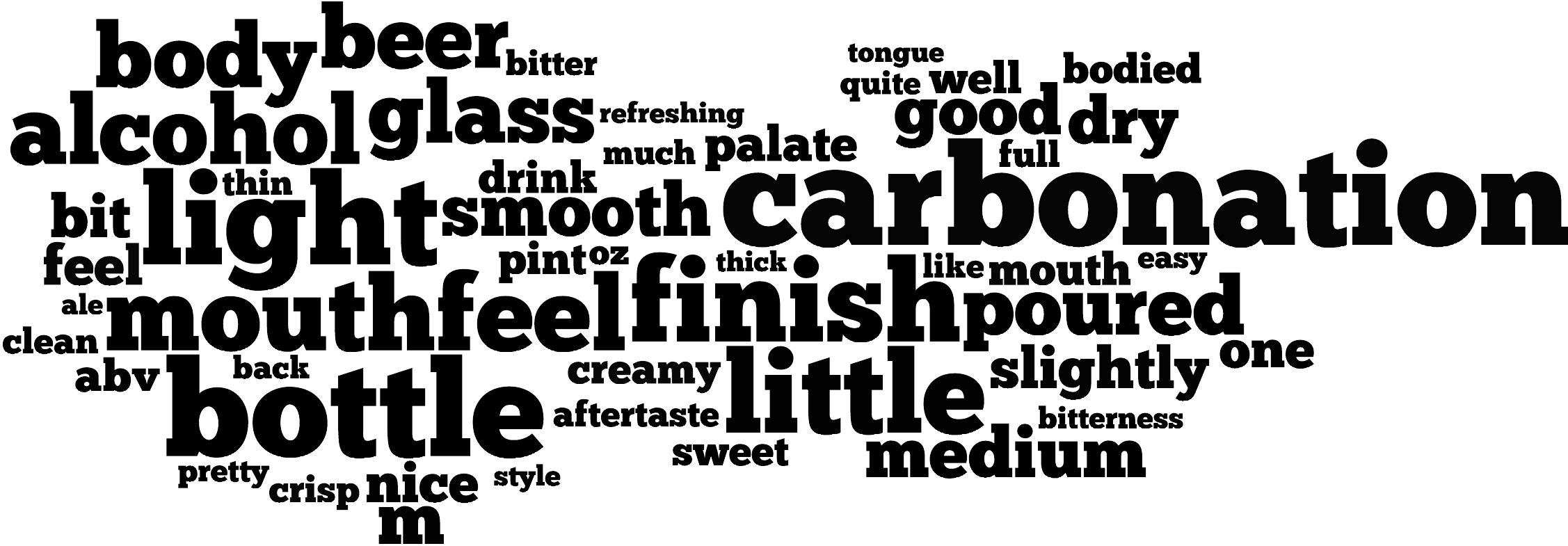}} & {\includegraphics[angle=-90,scale=0.2]{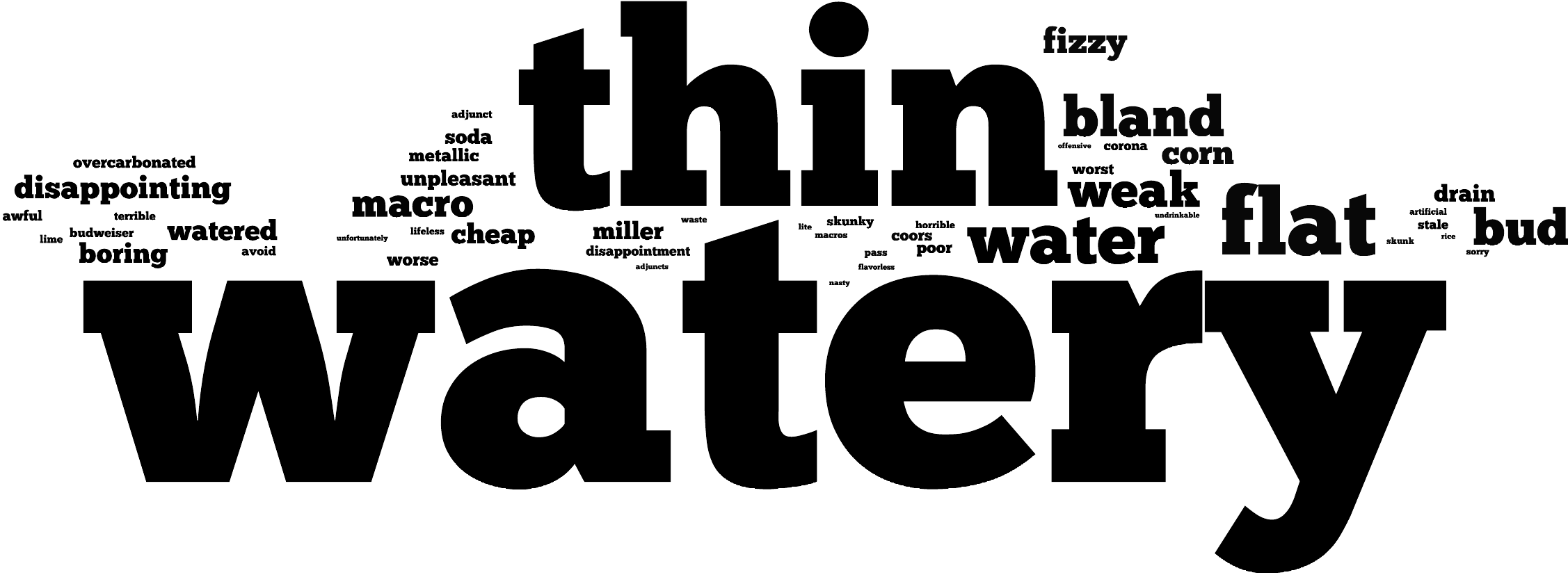}} & {\includegraphics[angle=-90,scale=0.2]{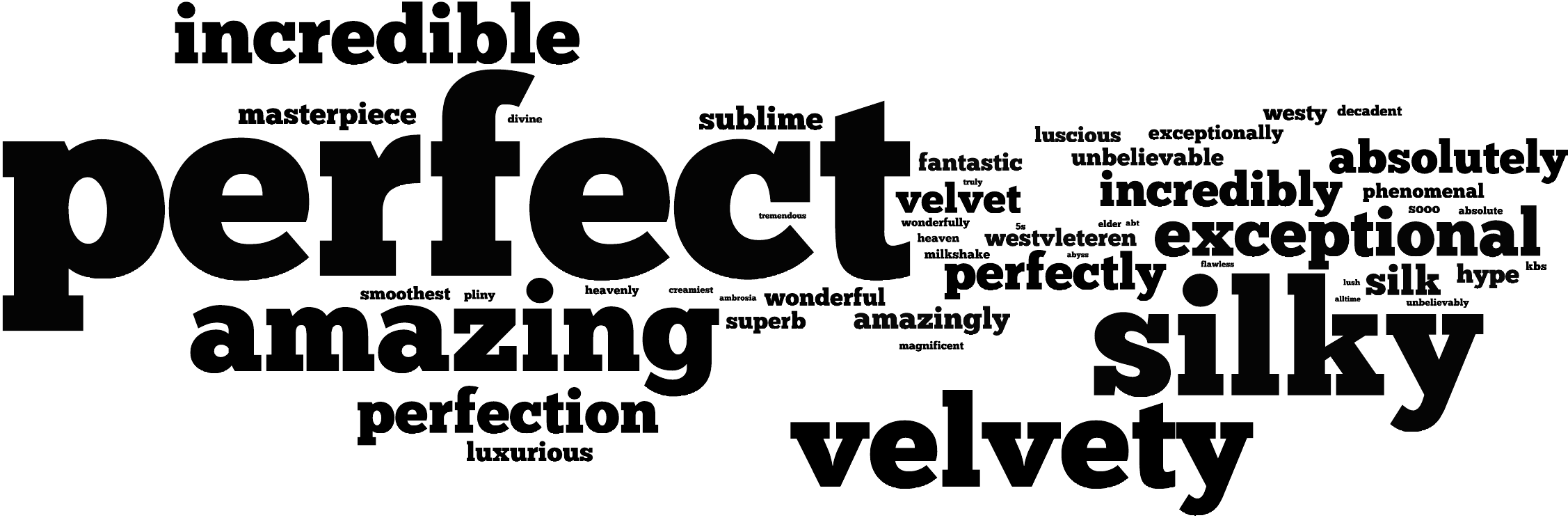}}\\[-3pt]
Look & {\includegraphics[angle=-90,scale=0.2]{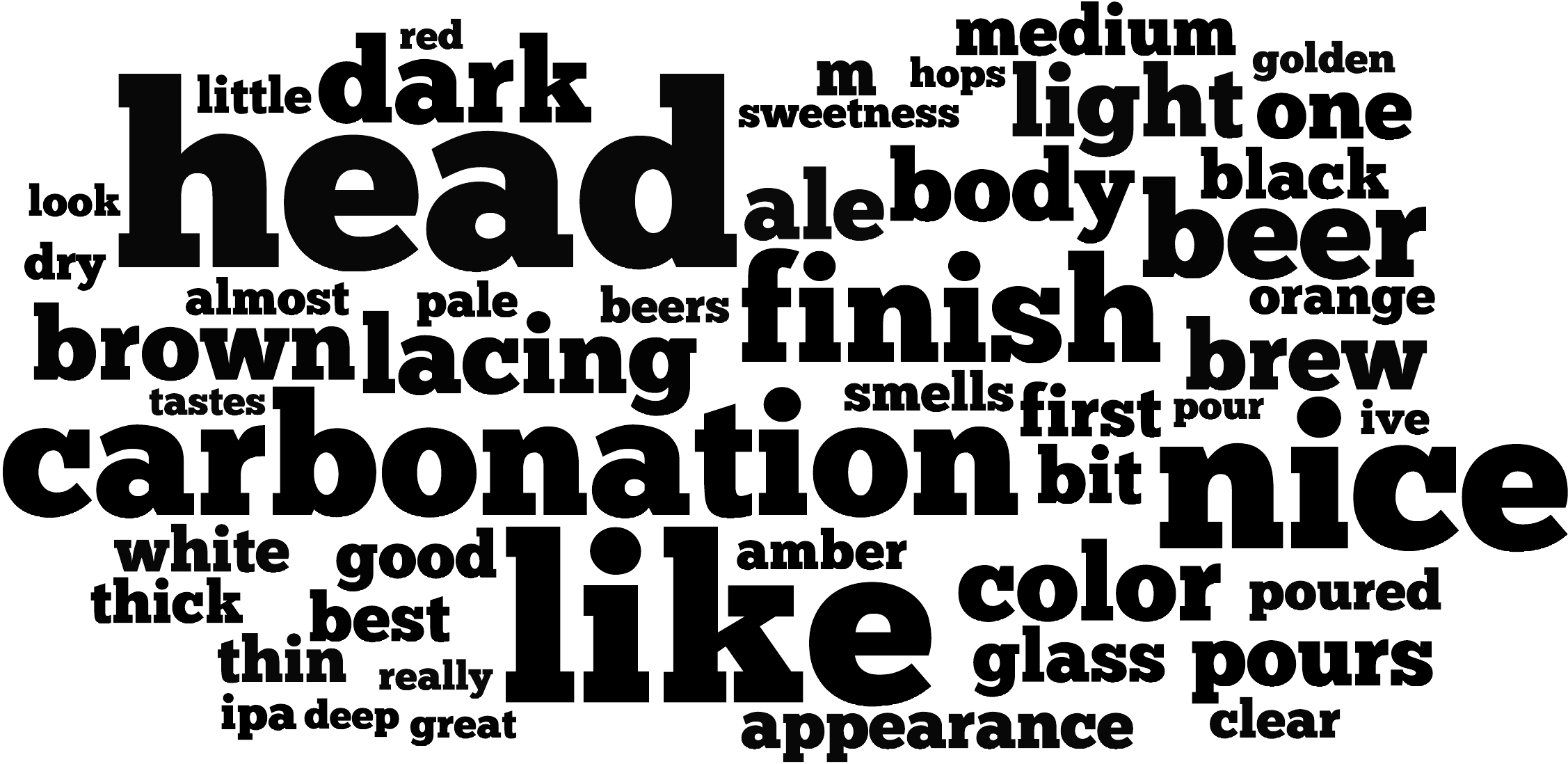}} & {\includegraphics[angle=-90,scale=0.2]{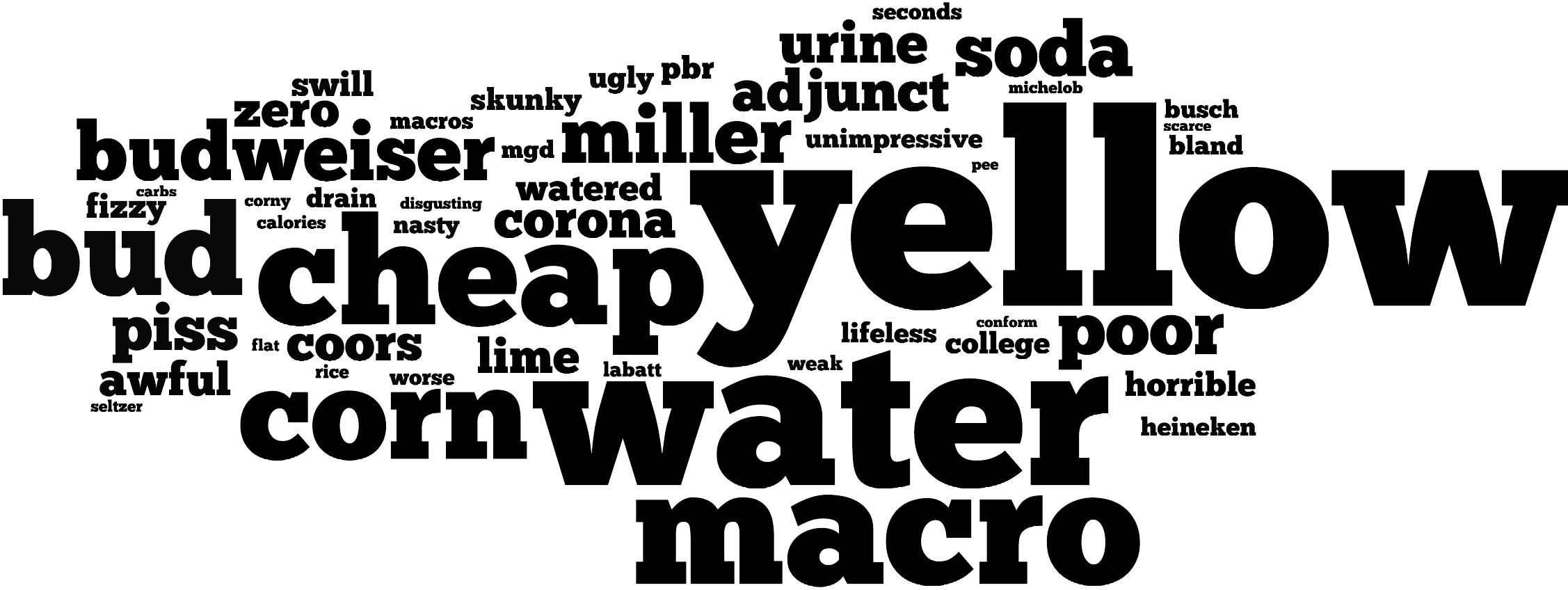}} & {\includegraphics[angle=-90,scale=0.2]{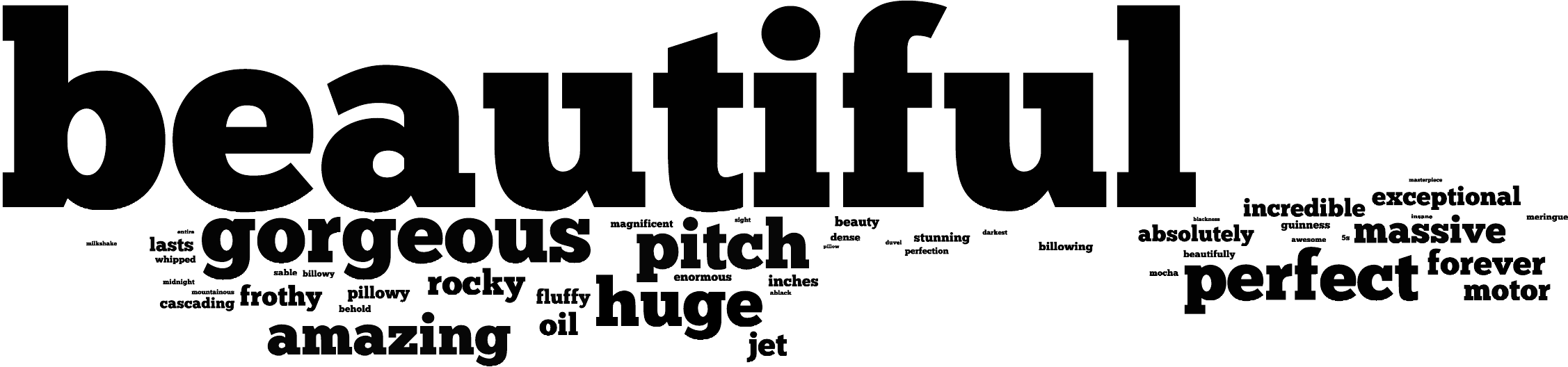}}\\[-3pt]
Smell & {\includegraphics[angle=-90,scale=0.2]{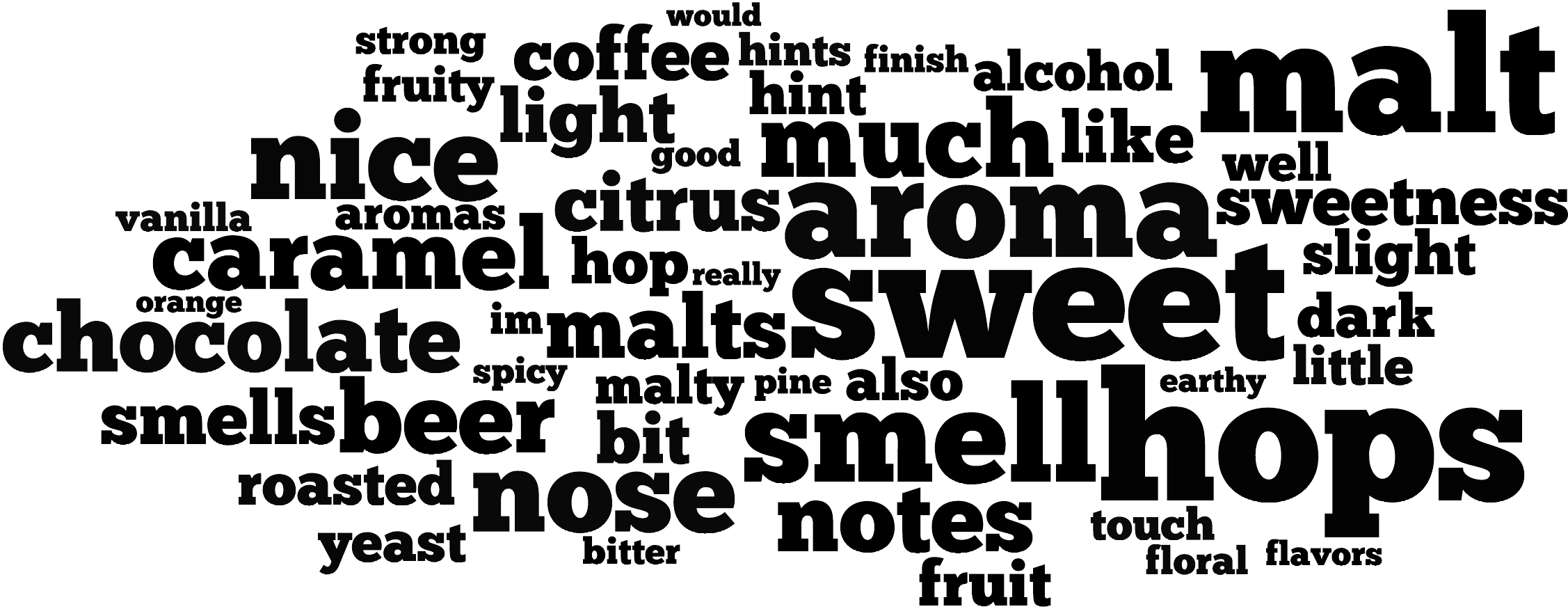}} & {\includegraphics[angle=-90,scale=0.2]{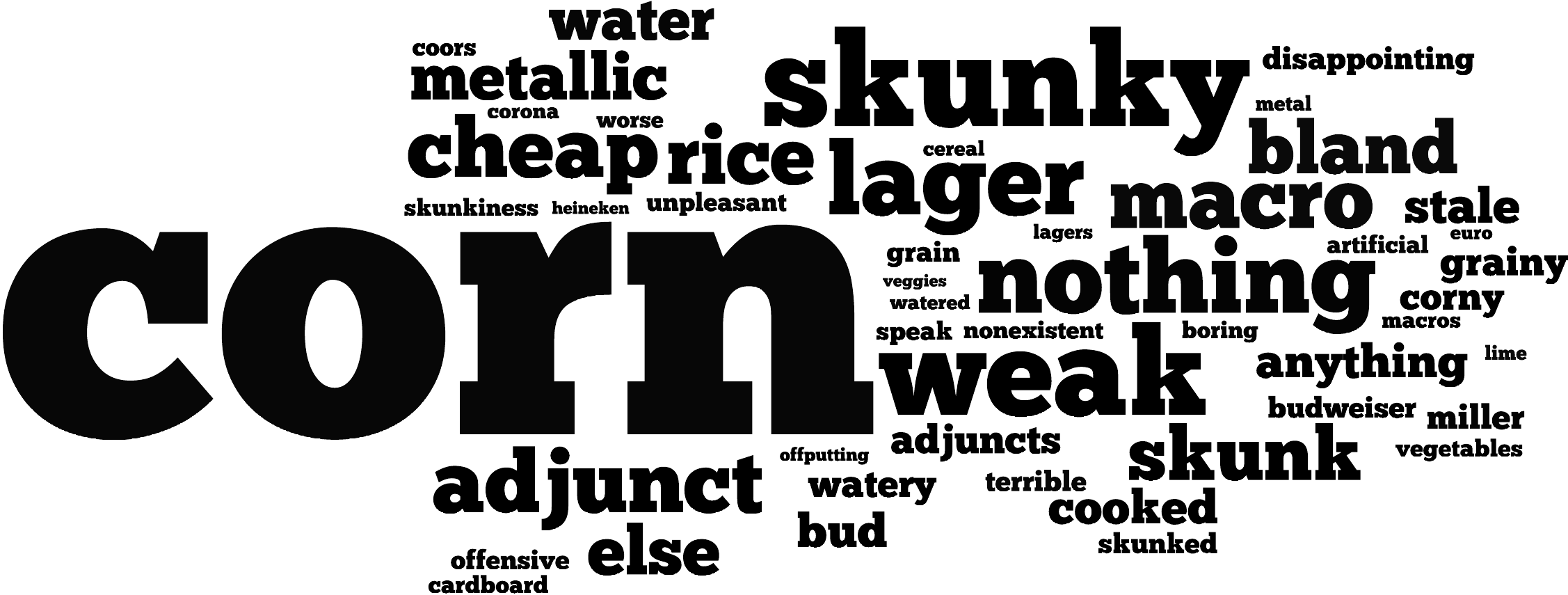}} & {\includegraphics[angle=-90,scale=0.2]{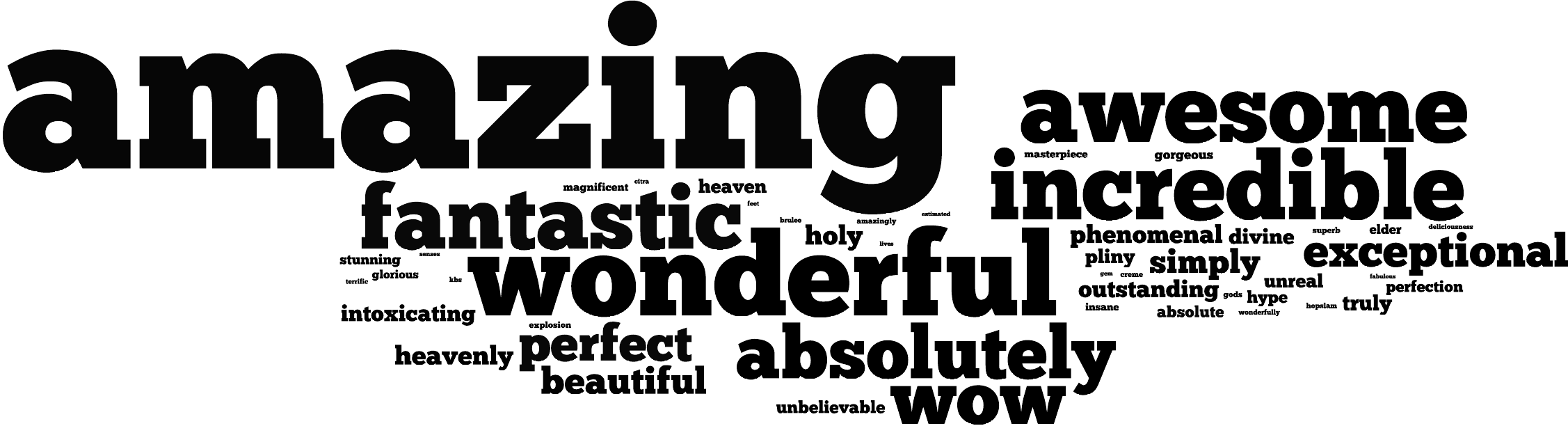}}\\[-3pt]
Taste & {\includegraphics[angle=-90,scale=0.2]{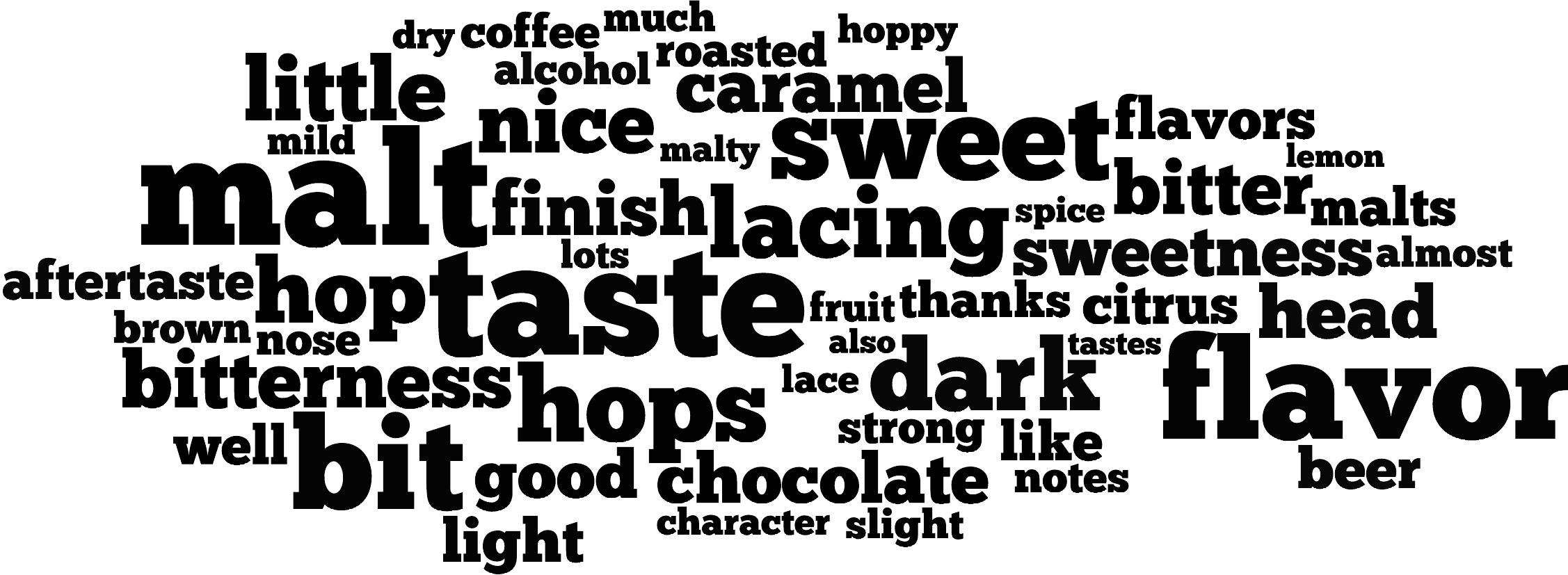}} & {\includegraphics[angle=-90,scale=0.2]{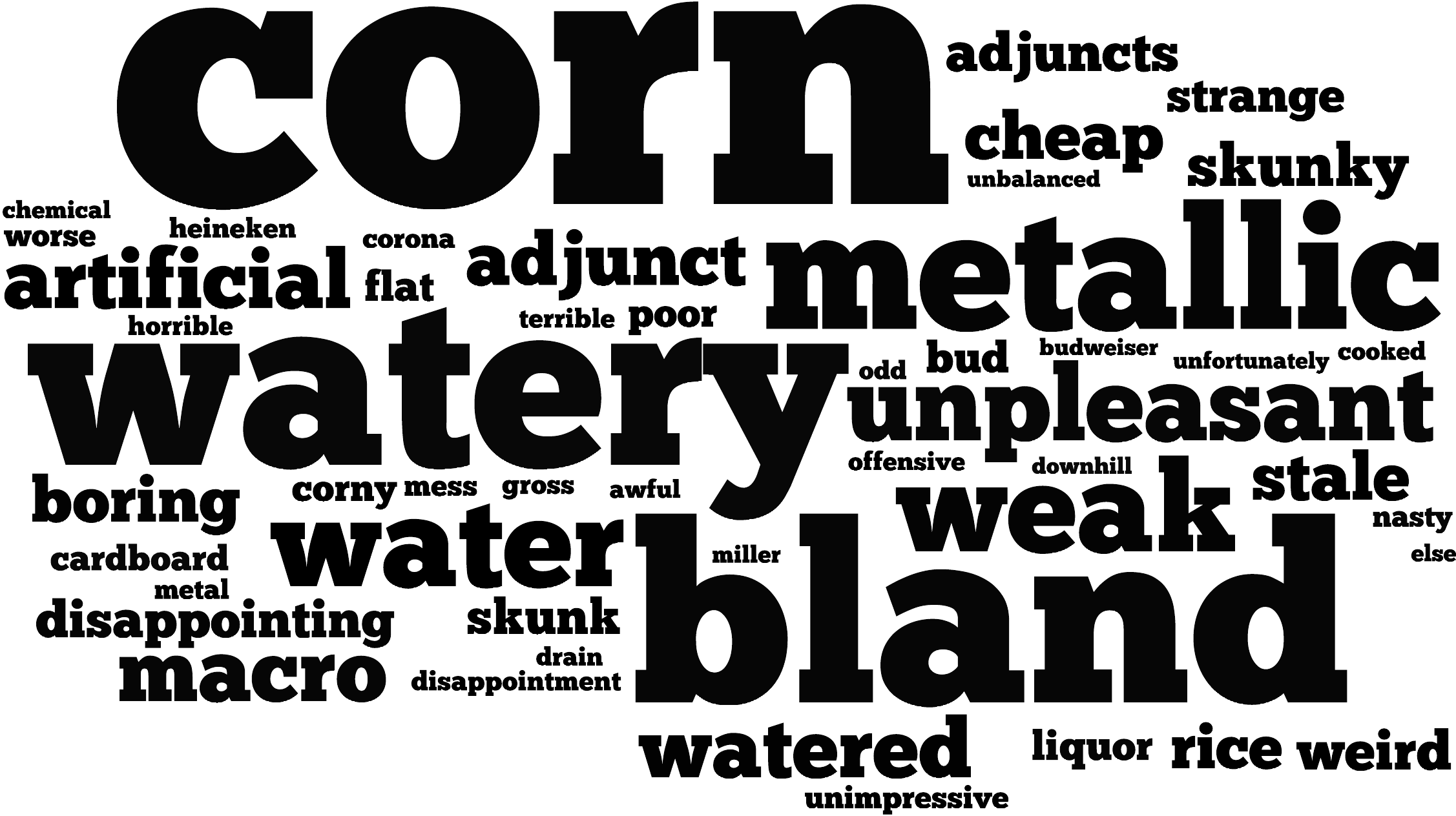}} & {\includegraphics[angle=-90,scale=0.2]{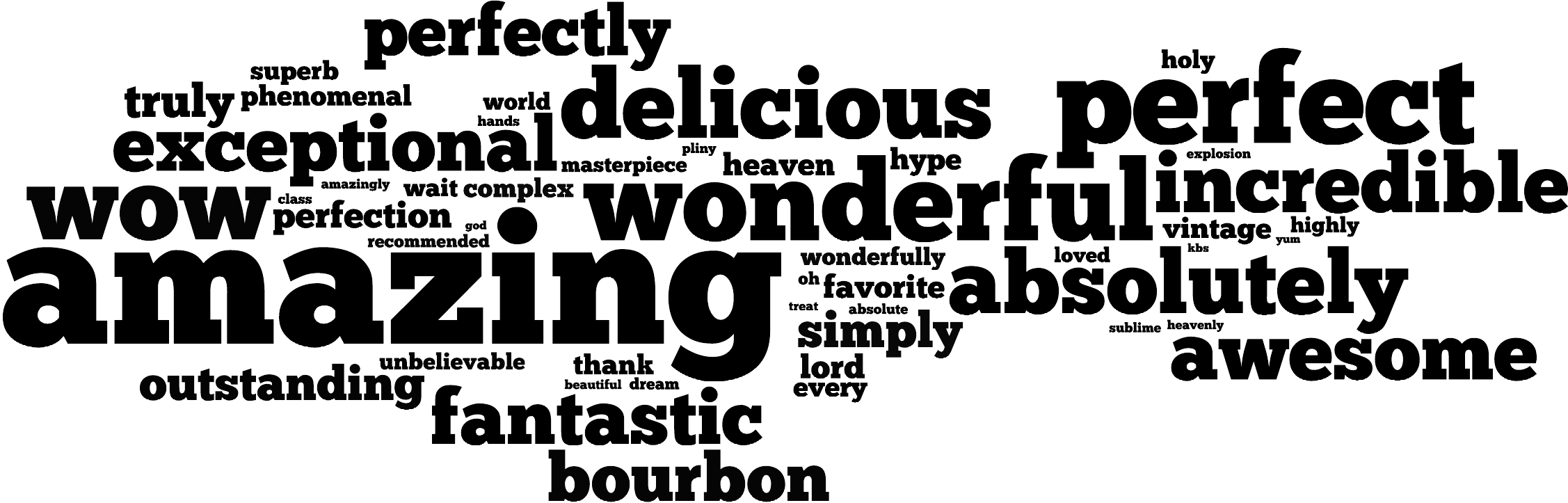}}\\[-6pt]
Overall & {\includegraphics[angle=-90,scale=0.2]{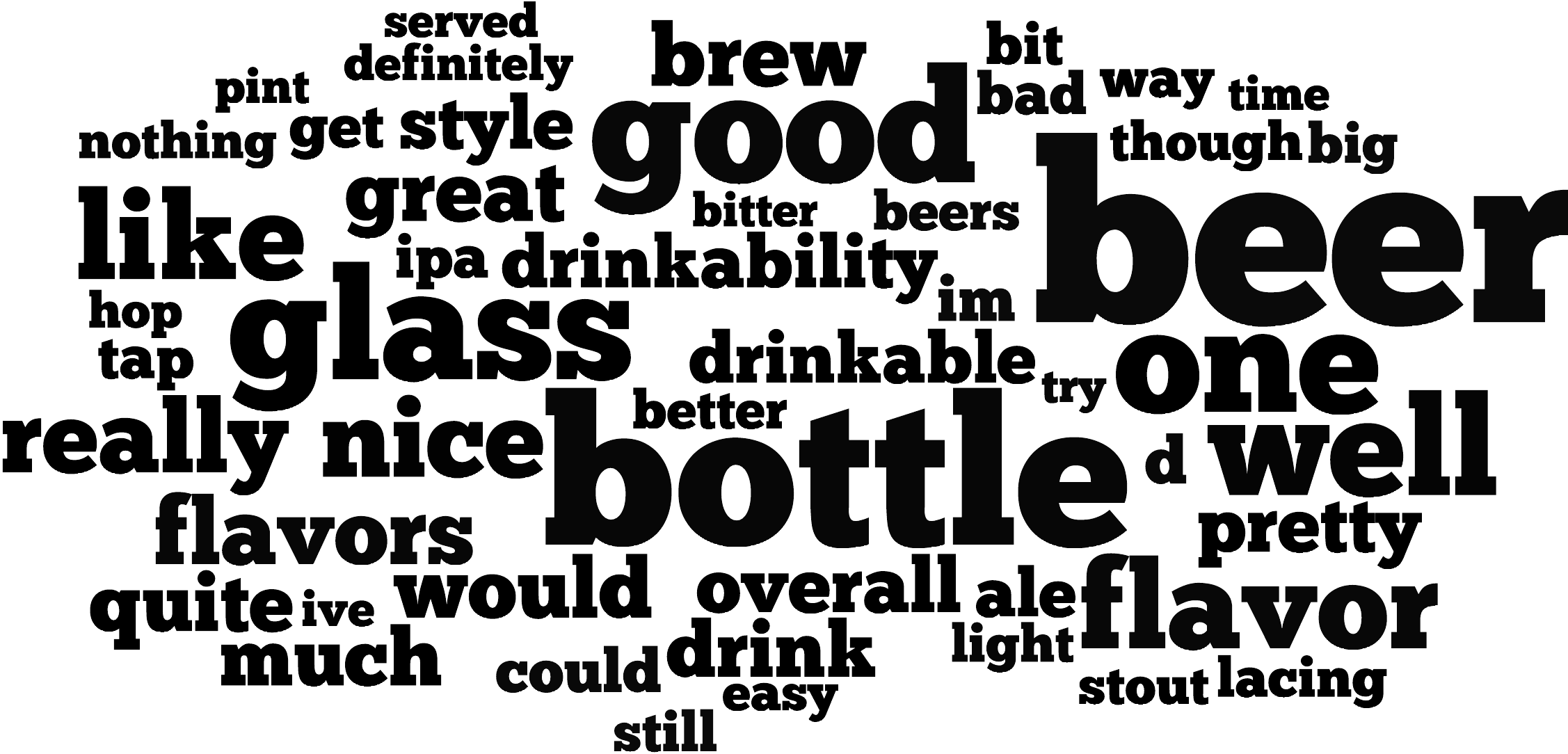}} & {\includegraphics[angle=-90,scale=0.2]{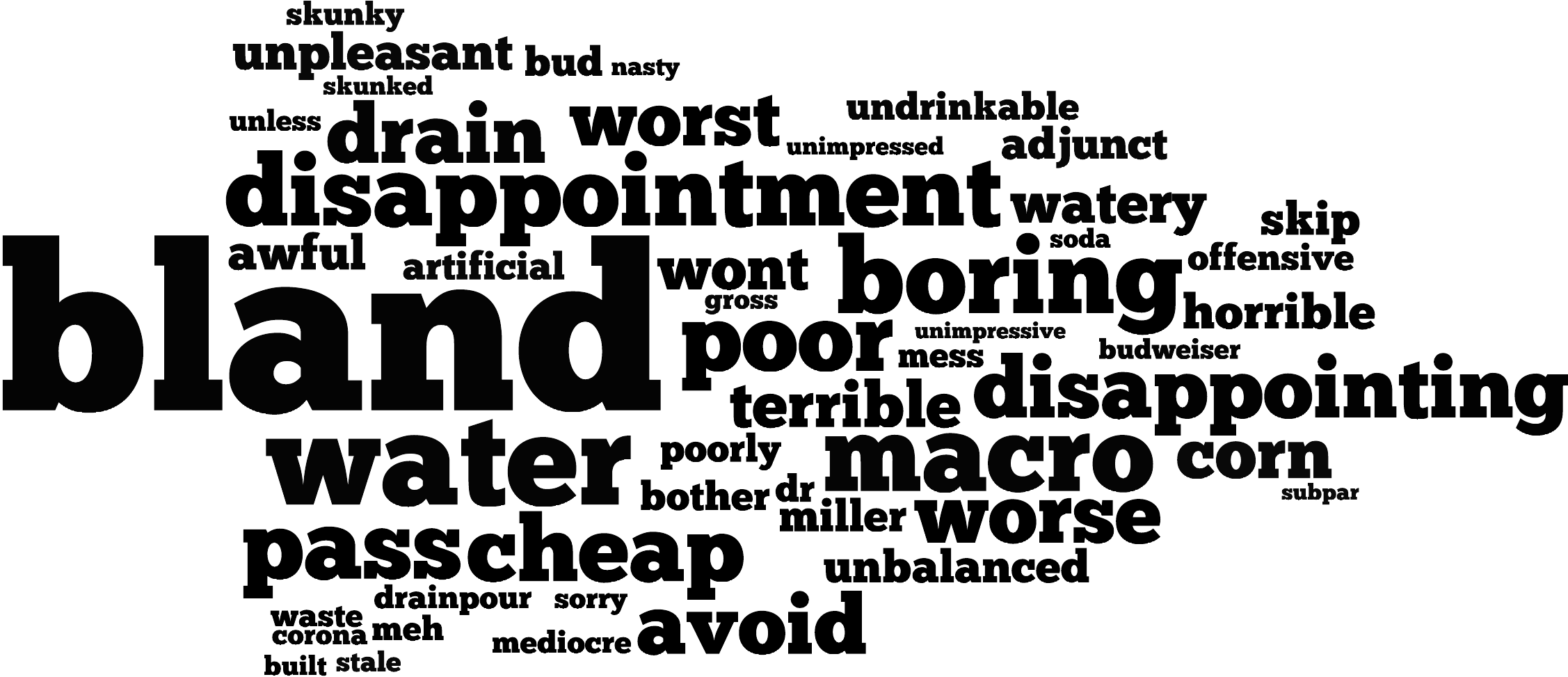}} & {\includegraphics[angle=-90,scale=0.2]{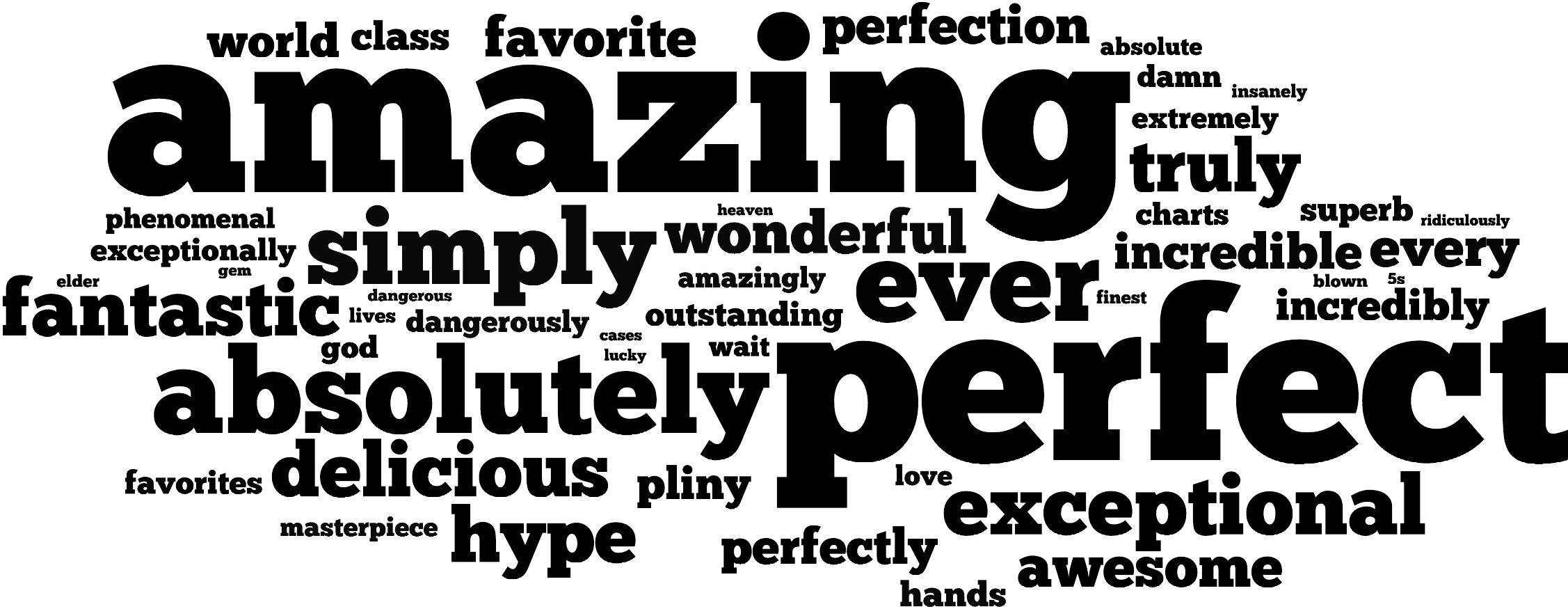}}
\end{tabular}
\caption{Word-cloud visualization of aspect parameters $\theta_{k}$ and sentiment parameters $\phi_{kv_k}$ learned from \emph{BeerAdvocate} data. Word sizes reflect the weights $\theta_{kw}$ and $\phi_{kv_kw}$ for each word $w$. Rows show different aspects $k$; the left column shows `aspect' weights $\theta_k$, the center column shows 2-star `sentiment' weights $\phi_{k,2}$, and the right column shows 5-star sentiment weights $\phi_{k,5}$ (1-star sentiment weights proved too unwholesome for publication). Parameters in this figure were learned using the \emph{unsupervised} version of our model. \label{fig:wordle}}
\end{figure*}

\subsection{Review Summarization}
\label{sec:summarization}

In the context of our model, summarization means identifying a subset of sentences that best explain a user's multiple-aspect rating.
Specifically, for each review $r_i$ and aspect $k$, we choose the sentence that maximizes the score for that aspect given the aspect's rating $v_{ik}$, as described in Section \ref{sec:learning}. This setup is motivated by the findings of \citet{lerman}; they show that users prefer summaries that discuss \emph{sentiments} about various aspects of a product, rather than merely the aspects themselves.

Results for this task are shown in Figure \ref{fig:perf} (middle). As before, increased supervision improves performance in almost all cases (semi-supervised learning beats unsupervised learning by 34\%, and fully-supervised learning further improves performance by 17\%). Note that summarization is not necessarily `easier' than segmentation, and both have higher scores on different datasets. Summarization is easiest when users discuss a variety of aspects, while segmentation is easiest when users primarily discuss `easy to classify' aspects. In practice, performance on both tasks is highly correlated. For this task, \modelname outperforms LDA significantly, since LDA incorrectly labels infrequently-discussed aspects, and doesn't make use of rating data.

\subsubsection{Aspect Ranking}
\label{sec:ranking}

Some works define summarization in terms of \emph{ranking} \citet{brody,titov,Snyder07multipleaspect}. For each aspect, probabilities are computed for each sentence, which are sorted to produce a ranking. Summarization can then be cast as retrieving the most relevant sentences for each aspect. Although the data from \citet{titov} are not available, for the sake of comparison we reproduce their experimental setup and baselines.

Figure \ref{fig:prcurves} shows aspect ranking results on \emph{BeerAdvocate} data. On their own data, \citet{titov} reported that their unsupervised method performed only 5\% worse than a fully-supervised maxent classifier. We report a similar result for our own unsupervised method (MAP=0.76 vs.~0.82), though we find that unsupervised learning outperforms maxent classification for two out of five aspects.
Furthermore, our semi-supervised algorithm matches the performance of maxent classification, and our fully-supervised method outperforms it by 7\%.

\subsection{Rating Prediction}

In many of the datasets we consider, only `overall' ratings are compulsory while aspect ratings are optional. In this section we try to recover such missing aspect ratings.
To measure performance on this task we train on half of our reviews to predict ratings for the other half. Naturally we ensure that none of the data used for evaluation were used during any stage of training, i.e., the segmentation models used in this experiment were \emph{not} trained using the reviews on which we predict ratings.

Rating prediction performance is shown in Figure \ref{fig:perf} (bottom). We exclude \emph{Pubs} data as it includes no overall rating, and \emph{CitySearch} data as ratings are per-sentence rather than per-review. As expected, ratings predicted from unsegmented text are inaccurate, as conflicting sentiments may appear for different aspects. More surprisingly, using \emph{segmented} text does \emph{not} solve this problem (in fact it is 32\% \emph{worse}), even when we have accurate aspect labels. A similar result was reported by \citep{lu11}, who found that models capable of segmenting text from ratings are not necessarily good at predicting ratings from text, and in fact such models do not outperform simple Support Vector Regression baselines. This occurs because aspect ratings are \emph{correlated}, and predicting ratings from segmented text fails to account for this correlation.

Our pairwise rating model, which explicitly models relationships between aspects, largely addresses this issue. Combining our rating model with unsegmented text already decreases the error by 23\% compared to the SVM baseline, and combining our rating model with \emph{segmented} text decreases the error by a further 22\%.

While segmented text improves upon unsegmented text, the level of supervision has little impact on performance for rating prediction. This is surprising, since supervision affects \emph{segmentation} performance significantly, and in some cases we obtain good performance on rating prediction even when aspect labels are inaccurate. To understand this, note that our unsupervised algorithm learns words that are highly correlated with users' ratings, which ultimately means that the labels it predicts must in some way be predictive of ratings, even if those labels are incorrect from the perspective of human annotators.
Pleasingly, this means that we can train a model to predict aspect ratings using an \emph{unsupervised} segmentation model; in other words good performance on our rating prediction task can be achieved without the intervention of human annotators.

\subsection{Qualitative Analysis}

We now examine the aspect and sentiment lexicons produced by our model. Figure \ref{fig:wordle} visualizes the learned parameters $\theta_k$ and $\phi_{kv_k}$ for the unsupervised version of our segmentation model on \emph{BeerAdvocate} data. We make a few observations: First, the weights match our intuition, e.g.~words like `carbonation', `head', `aroma', and `flavor' match the aspects to which they are assigned. Second, the highest weighted words for `aspect' parameters are predominantly \emph{nouns}, while the highest weighted words for `sentiment' parameters are predominantly \emph{adjectives}; this confirms that aspect and sentiment words fulfil their expected roles. Third, we find that very different words are used to describe different sentiments, e.g.~`watery' has high weight for feel and taste, but not for look and smell; the study `Old Wine or Warm Beer' \citep{oldwine} discusses how nouns and adjectives interact in this type of data, supporting our decision to model `aspect' and `sentiment' words separately, and to include separate sentiment parameters for each aspect.

To explain why `corn' is a 2-star smell and taste word, note that corn is not normally an ingredient of beer. It is used in place of barley in inexpensive, mass-produced beers (so called `adjuncts'), which account for many of the 1- and 2-star reviews in our corpus; thus it is not surprising that the word has negative connotations among beer enthusiasts.

%% file: 080conclusion.tex
\section{Conclusion}

By introducing corpora of five million reviews from five sources, we have studied review systems in which users provide ratings for multiple \emph{aspects} of each product. By learning which words describe each aspect and the associated sentiment, our model is able to determine which parts of a review correspond to each rated aspect, which sentences best summarize a review, and how to recover ratings that are missing from reviews. We learn highly interpretable aspect and sentiment lexicons, and our model readily scales to the real-world corpora we consider.

%% file: arxiv.bbl
\small
\begin{thebibliography}{10}

\bibitem{baccianella09}
S.~Baccianella, A.~Esuli, and F.~Sebastiani.
\newblock Multi-facet rating of product reviews.
\newblock In {\em ECIR}, 2009.

\bibitem{netflix}
J.~Bennett and S.~Lanning.
\newblock The {Netflix} prize.
\newblock In {\em KDD Cup and Workshop}, 2007.

\bibitem{blair08}
S.~Blair-Goldensohn, T.~Neylon, K.~Hannan, G.~Reis, R.~McDonald, and J.~Reynar.
\newblock Building a sentiment summarizer for local service reviews.
\newblock In {\em NLP in the Information Explosion Era}, 2008.

\bibitem{blei}
D.~Blei, A.~Ng, and M.~Jordan.
\newblock Latent dirichlet allocation.
\newblock {\em JMLR}, 2003.

\bibitem{brody}
S.~Brody and N.~Elhadad.
\newblock An unsupervised aspect-sentiment model for online reviews.
\newblock In {\em ACL}, 2010.

\bibitem{lgm}
T.~Caetano, J.~McAuley, L.~Cheng, Q.~Le, and A.~Smola.
\newblock Learning graph matching.
\newblock {\em PAMI}, 2009.

\bibitem{cohen}
J.~Cohen.
\newblock A coefficient of agreement for nominal scales.
\newblock {\em Edu. and Psych. Measurement}, 1960.

\bibitem{du10}
L.~Du, W.~Buntine, and H.~Jin.
\newblock A segmented topic model based on the two-parameter poisson-dirichlet
  process.
\newblock {\em Machine Learning}, 2010.

\bibitem{ErkanRadev04}
G.~Erkan and D.~Radev.
\newblock Lexrank: Graph-based centrality as salience in text summarization.
\newblock {\em JAIR}, 2004.

\bibitem{oldwine}
A.~Fahrni and M.~Klenner.
\newblock Old wine or warm beer: Target-specific sentiment analysis of
  adjectives.
\newblock In {\em Affective Language in Human and Machine}, 2008.

\bibitem{Gamon05}
M.~Gamon, A.~Aue, S.~Corston-Oliver, and E.~Ringger.
\newblock Pulse: Mining customer opinions from free text.
\newblock In {\em IDA}, 2005.

\bibitem{ganu}
G.~Ganu, N.~Elhadad, and A.~Marian.
\newblock Beyond the stars: Improving rating predictions using review text
  content.
\newblock In {\em WebDB}, 2009.

\bibitem{gupta10}
N.~Gupta, G.~Di~Fabbrizio, and P.~Haffner.
\newblock Capturing the stars: predicting ratings for service and product
  reviews.
\newblock In {\em HLT Workshops}, 2010.

\bibitem{Hatz98}
V.~Hatzivassiloglou and K.~McKeown.
\newblock Predicting the semantic orientation of adjectives.
\newblock In {\em ACL}, 1997.

\bibitem{hoffman}
M.~Hoffman, D.~Blei, and F.~Bach.
\newblock Online learning for latent dirichlet allocation.
\newblock In {\em NIPS}, 2010.

\bibitem{hu04}
M.~Hu and B.~Liu.
\newblock Mining and summarizing customer reviews.
\newblock In {\em KDD}, 2004.

\bibitem{jo11}
Y.~Jo and A.~Oh.
\newblock Aspect and sentiment unification model for online review analysis.
\newblock In {\em WSDM}, 2011.

\bibitem{koren}
Y.~Koren, R.~Bell, and C.~Volinsky.
\newblock Matrix factorization techniques for recommender systems.
\newblock {\em Computer}, 2009.

\bibitem{cameras}
H.~Lakkaraju, C.~Bhattacharyya, I.~Bhattacharya, and S.~Merugu.
\newblock Exploiting coherence for the simultaneous discovery of latent facets
  and associated sentiments.
\newblock In {\em SDM}, 2011.

\bibitem{lerman}
K.~Lerman, S.~Blair-Goldensohn, and R.~McDonald.
\newblock Sentiment summarization: evaluating and learning user preferences.
\newblock In {\em ACL}, 2009.

\bibitem{lin09}
C.~Lin and Y.~He.
\newblock Joint sentiment/topic model for sentiment analysis.
\newblock In {\em CIKM}, 2009.

\bibitem{lu11}
B.~Lu, M.~Ott, C.~Cardie, and B.~Tsou.
\newblock Multi-aspect sentiment analysis with topic models.
\newblock In {\em Workshop on SENTIRE}, 2011.

\bibitem{lu09}
Y.~Lu, C.~Zhai, and N.~Sundaresan.
\newblock Rated aspect summarization of short comments.
\newblock In {\em WWW}, 2009.

\bibitem{icdm}
J.~McAuley, J.~Leskvec, and D.~Jurafsky.
\newblock Learning attitudes and attributes from multi-aspect reviews.
\newblock In {\em ICDM}, 2012.

\bibitem{mohammad09}
S.~Mohammad, C.~Dunne, and B.~Dorr.
\newblock Generating high-coverage semantic orientation lexicons from overtly
  marked words and a thesaurus.
\newblock In {\em EMNLP}, 2009.

\bibitem{ng06}
R.~Ng and A.~Pauls.
\newblock Multi-document summarization of evaluative text.
\newblock In {\em ACL}, 2006.

\bibitem{popescu05}
A.~Popescu and O.~Etzioni.
\newblock Extracting product features and opinions from reviews.
\newblock In {\em HLT}, 2005.

\bibitem{rao09}
D.~Rao and D.~Ravichandran.
\newblock Semi-supervised polarity lexicon induction.
\newblock In {\em ACL}, 2009.

\bibitem{Snyder07multipleaspect}
B.~Snyder and R.~Barzilay.
\newblock Multiple aspect ranking using the good grief algorithm.
\newblock In {\em ACL}, 2007.

\bibitem{titov}
I.~Titov and R.~McDonald.
\newblock A joint model of text and aspect ratings for sentiment summarization.
\newblock In {\em ACL}, 2008.

\bibitem{titovMultigrain}
I.~Titov and R.~McDonald.
\newblock Modeling online reviews with multi-grain topic models.
\newblock In {\em WWW}, 2008.

\bibitem{tsoch05}
I.~Tsochantaridis, T.~Joachims, T.~Hofmann, and Y.~Altun.
\newblock Large margin methods for structured and interdependent output
  variables.
\newblock {\em JMLR}, 2005.

\bibitem{turney02}
P.~Turney.
\newblock Thumbs up or thumbs down? semantic orientation applied to
  unsupervised classification of reviews.
\newblock In {\em ACL}, 2002.

\bibitem{velikovich10}
L.~Velikovich, S.~Blair-Goldensohn, K.~Hannan, and R.~McDonald.
\newblock The viability of web-derived polarity lexicons.
\newblock In {\em HLT}, 2010.

\bibitem{latentSVM}
C.-N. Yu and T.~Joachims.
\newblock Learning structural {SVM}s with latent variables.
\newblock In {\em ICML}, 2009.

\bibitem{zhao10}
W.~Zhao, J.~Jiang, H.~Yan, and X.~Li.
\newblock Jointly modeling aspects and opinions with a {MaxEnt-LDA} hybrid.
\newblock In {\em EMNLP}, 2010.

\end{thebibliography}
